%% file: neurips_2026.tex
\documentclass{article}

\usepackage[preprint]{neurips_2026}
\usepackage[T1]{fontenc}    
\usepackage{hyperref}       
\usepackage{url}            
\usepackage{booktabs}       
\usepackage{amsfonts}       
\usepackage{nicefrac}       
\usepackage{microtype}      
\usepackage{xcolor}         
\usepackage{listings}
\usepackage{xspace}
\usepackage{enumitem}
\usepackage{amsmath}
\usepackage{cleveref}
\usepackage{graphics}
\usepackage{graphicx}
\usepackage{multirow}

\makeatletter
\AddToHook{cmd/appendix/before}{\def\cref@section@alias{appendix}}
\makeatother

\usepackage[most]{tcolorbox}
\newtcolorbox{defbox}[1]{
  enhanced,
  breakable,
  colback=gray!3,
  colframe=black!55,
  boxrule=0.4pt,
  arc=2pt,
  left=10pt, right=10pt, top=6pt, bottom=6pt,
  fonttitle=\bfseries,
  coltitle=black,
  title=#1,
}

\definecolor{keywordcolor}{rgb}{0.7, 0.1, 0.1}   
\definecolor{tacticcolor}{rgb}{0.0, 0.1, 0.6}    
\definecolor{commentcolor}{rgb}{0.4, 0.4, 0.4}   
\definecolor{symbolcolor}{rgb}{0.0, 0.1, 0.6}    
\definecolor{sortcolor}{rgb}{0.1, 0.5, 0.1}      
\definecolor{attributecolor}{rgb}{0.7, 0.1, 0.1} 

\newcommand{\code}[1]{\lstinline|#1|}

\newcommand{\modeproof}{proof-only\xspace}
\newcommand{\modecodeproof}{code-and-proof\xspace}

\newif\ifcomments
\commentstrue
\ifcomments
\newcommand{\todo}[1]{\textcolor{orange}{[TODO: #1]}}
\newcommand{\zhe}[1]{\textcolor{blue}{[zhe: #1]}}
\newcommand{\jingxuan}[1]{\textcolor{teal}{[jingxuan: #1]}}
\newcommand{\dawn}[1]{\textcolor{red}{[dawn: #1]}}
\else
\newcommand{\todo}[1]{}
\newcommand{\zhe}[1]{}
\newcommand{\jingxuan}[1]{}
\newcommand{\dawn}[1]{}
\fi

\newcommand{\benchmark}{Vero\xspace}
\newcommand{\repoimpl}{\texttt{RepoImpl}}
\newcommand{\canonical}{\texttt{canonical}}

\newcommand{\smallsec}[1]{\textbf{#1}}

\title{\benchmark: Can AI Agents Build Formally Verified Software Repositories?
}

\author{%
Zhe Ye\textsuperscript{1*}
\And
Hantao Lou\textsuperscript{1*}
\And
Yuechun Sun\textsuperscript{2*}
\And
Peiyang Song\textsuperscript{3}
\And
Zhengxu Yan\textsuperscript{4}
\And
Timothe Kasriel\textsuperscript{1}
\AND
Qingyang Zhang\textsuperscript{1}
\And
Kaiyu Yang\textsuperscript{5}
\And
Soonho Kong\textsuperscript{6\dag}
\And
Jingxuan He\textsuperscript{1}
\And
Dawn Song\textsuperscript{1}
\AND\normalfont
\textsuperscript{1}UC Berkeley \quad
\textsuperscript{2}University of Chicago \quad
\textsuperscript{3}California Institute of Technology \\
\textsuperscript{4}Stanford University \quad
\textsuperscript{5}Apodex \quad
\textsuperscript{6}Amazon Web Services \\
}

\begin{document}

\maketitle
\let\thefootnote\relax\footnotetext{\textsuperscript{*}Equal contribution.}\footnotetext{\textsuperscript{\dag}Work does not relate to position at Amazon.}

\begin{abstract}

AI agents are increasingly used for programming, but do not provide any guarantee on the correctness of generated code.
Verified code generation, in which an agent produces both an implementation and a machine-checked proof of its specification, offers a stronger path toward trustworthy AI-generated software.
Existing benchmarks in this direction either focus on individual functions or only evaluate proof generation with provided implementations.
It is still an open question whether agents can make coherent implementation and proof choices across real multi-module codebases.
To bridge this gap, we introduce \benchmark{}, the first benchmark to evaluate joint implementation and proof synthesis at the repository level.
\benchmark{} contains 43 multi-module instances sourced from real-world repositories spanning Python, Dafny, Verus, and Coq, and covering diverse domains from cryptographic protocols to distributed systems.
Each instance consists of a multi-module Lean 4 repository with predetermined API interfaces, manually curated formal specifications, and reference implementations, supporting both proof-only and code-and-proof evaluation modes.
To improve benchmark reliability, \benchmark{} also includes an audit mechanism where agents are allowed to formally prove unsatisfiability of provided specification or incorrectness of reference code, which surfaces and corrects latent code and specification errors during curation.
We evaluate frontier coding-agent configurations with Lean toolchain access.
The strongest agent fully solves only 27 of 43 instances and closes no specifications on the hardest repositories.
\benchmark{} provides a concrete testbed for measuring progress toward repository-scale verified software synthesis, where current agents still fall short.
We release the benchmark, curation pipeline, and evaluation harness at {\color{blue} \url{https://github.com/sunblaze-ucb/vero}}.

\end{abstract}

\section{Introduction}
\label{sec:intro}






AI agents are increasingly used in software engineering tasks~\citep{jimenez2024swe,yang2024swe}.
The correctness of agent-produced code, however, is typically assessed through unit tests and human review.
These methods catch anticipated failures but cannot rule out edge-case bugs and security vulnerabilities, which are most damaging in protocol and infrastructure software.
Formal verification offers a stronger alternative: a machine-checked proof that an implementation satisfies its specification rules out, for all inputs, every class of bug the specification captures.
Therefore, it is critical to rigorously assess whether AI agents can perform formal verification at the scale of real-world software.

Existing benchmarks for verified code generation either target individual functions~\citep{lohn2024minicodeprops, sun2024clover, dougherty2025proving, ye2025verina, thakur2025clever, bursuc2025benchmark} or evaluate only proof generation with a fixed implementation~\citep{zhong2025towards, xin2026verisoftbench, yang2025verusage, rao2026s2n}, leaving the joint code-and-proof task unaddressed.
Repository-scale verified code generation does not reduce to scaling up function-level techniques, because code, specifications, and proofs are deeply interdependent across files: a proof for one function may depend on lemmas about lower-level utilities, and revising an implementation can invalidate proofs throughout the repository.
Agents must reason globally about the consistency of the entire codebase, rather than locally about a single function.
This is also where verified code generation matters in practice, since real-world verified software, from operating system kernels~\citep{klein2009sel4,gu2016certikos} to cryptographic protocols~\citep{polubelova2020haclxn,Erbsen2019SimpleHC} and distributed systems~\citep{Hawblitzel2015IronFleetPP}, lives in repositories rather than standalone functions.
Closing this evaluation gap is essential for measuring and driving progress on agents that can produce verified software at the scale for practical impact.

To bridge this gap, we introduce \benchmark, the first repository-level benchmark for verified code generation in Lean~4~\cite{de2015lean}.
\benchmark consists of 43 benchmark instances curated from real-world repositories spanning a diverse set of source languages, including Python, Dafny~\citep{leino2010dafny}, Verus~\citep{lattuada2023verus}, and Coq~\citep{barras1997coq}, and covering a range of domains and difficulty levels from foundational data structures to verified systems.
Sourcing from verification-aware languages allows us to leverage existing formal specifications and proofs as ground truth, while Python repositories expand domain coverage to widely used algorithmic and systems code.
Each instance is a Lean~4 project comprising multiple modules with their data type definitions, auxiliary definitions, \emph{API signatures}, and human-curated ground-truth \emph{specifications}.
In total, \benchmark{} contains 743 scored APIs and 2{,}705 specifications, so the 43 repositories correspond to thousands of verification obligations rather than 43 standalone tasks.
We constructed benchmark instances through a multi-stage curation pipeline where every translated definition, specification, and proof obligation is manually reviewed and validated by the authors.
The agent's task is to fit into this scaffold by discharging two types of obligations: an \emph{implementation obligation} for each API signature, and a \emph{proof obligation} for each specification.
\benchmark supports two task modes.
In the primary \emph{code-and-proof} mode, agents must discharge both obligations: synthesize implementations for every API and machine-checked Lean~4 proofs that they satisfy all specifications.
In the \emph{proof-only} mode, reference implementations are additionally provided and the agent discharges only the proof obligations.
\benchmark{} is, to our knowledge, the first benchmark that evaluates agents on joint code-and-proof generation at the repository level.

A persistent challenge in constructing formal-verification benchmarks is that even carefully curated specifications and reference implementations can contain subtle errors, a problem documented across formal verification projects and benchmarks \citep{feng2026certified}.
Such errors would silently penalize otherwise capable agents and treat benchmark defects as agent failures.
\benchmark addresses this through a novel audit mechanism.
Rather than treating such errors as unexplained agent failures, the mechanism accepts formal proofs of specification unsatisfiability or reference implementation incorrectness to flag the affected instance for curator review, and uses the formal witnesses to guide correction.
During curation, the audit mechanism surfaced several latent errors that escaped manual review, and helped guide the benchmark improvement by providing formal evidences like counter examples.

We evaluate four frontier model configurations under two coding-agent harnesses on \benchmark{} across both task modes.
\benchmark{} is frontier-resistant: the best-performing configuration fully solves only 27 of 43 instances in \modecodeproof, and 10 instances resist every configuration in both modes.
Our analysis further reveals that the unsolved instances concentrate on specifications encoding cross-module invariants, protocol consistency, and custom mathematical theories, where current agents attempt local proofs of individual specifications rather than building the reusable lemma libraries that verification at this scale requires.
We further find that implementation freedom in \modecodeproof cuts both ways.
Agents can sidestep hard-to-verify reference algorithms by writing simpler implementations that satisfy the same specifications.
On harder instances, however, agents fail to coordinate the two artifacts, producing rewrites that break existing proofs or introduce build errors elsewhere in the repository.
Together these findings provide concrete directions for advancing repository-scale verified code generation.

\smallsec{Contributions.}
In summary, we make the following key contributions:

\begin{itemize}[leftmargin=*]
    \item We introduce \benchmark, the first repository-scale benchmark for verified code generation in Lean~4, comprising 43 instances curated from real-world repositories across diverse source languages and domains, together with a semi-automated, multi-language curation pipeline that supports its construction and future extension.
    \item We design a formal audit mechanism that accepts machine-checked proofs of specification unsatisfiability or reference implementation incorrectness, turning latent benchmark errors into actionable findings and enabling continuous quality improvement as agents grow more capable.
    \item We conduct a comprehensive evaluation of state-of-the-art coding agents and frontier LLMs on \benchmark across two task modes, providing rigorous baselines and actionable insights toward scalable, automated software verification.
\end{itemize}

\section{Background and Related Work}
\label{sec:background}



Existing benchmarks for verified code generation share several limitations that constrain their utility for evaluating coding agents in realistic settings.

\smallsec{First, most benchmarks evaluate verified code generation only at the level of individual functions.}
In Lean~4, miniCodeProps~\citep{lohn2024minicodeprops}, FVAPPS~\citep{dougherty2025proving}, VERINA~\citep{ye2025verina}, and CLEVER~\citep{thakur2025clever} all target standalone tasks drawn from algorithmic problem sets such as HumanEval~\citep{chen2021evaluating}, MBPP~\citep{austin2021program}, and APPS~\citep{hendrycks2021measuringcoding}.
The same pattern holds in benchmarks for SMT-based verifiers such as Dafny~\citep{leino2010dafny} and Verus~\citep{lattuada2023verus}: DafnyBench~\citep{loughridge2024dafnybench}, VerusBench~\citep{yang2025autoverus}, AlgoVeri~\citep{zhao2026algoveri}, VeriCoding~\citep{bursuc2025benchmark}, VerifyThisBench~\citep{deng2025verifythisbench}, and VeriEquivBench~\citep{zeng2025veriequivbench} all operate on individual algorithms or methods.
While these benchmarks have driven measurable progress on isolated tasks, they cannot capture the cross-module dependencies and long-horizon reasoning that real-world verified codebases require.
\benchmark{} addresses this gap by curating multi-module Lean~4 projects with type definitions, API signatures, and specifications spread across files, requiring agents to reason about the consistency of the entire repository.

\smallsec{Second, the few existing repository-scale benchmarks evaluate proof generation alone, leaving the joint code-and-proof task unaddressed.}
RVBench~\citep{zhong2025towards} provides 755 proof completion tasks across four Verus projects; VeruSAGE-Bench~\citep{yang2025verusage} contains 849 proof tasks extracted from eight Verus-verified Rust systems; and VeriSoftBench~\citep{xin2026verisoftbench} packages 500 Lean~4 proof obligations from open-source formal-methods developments.
For Coq, CoqStoq~\citep{thompson2024rango} offers a large corpus of repository-level theorems for proof synthesis, while case studies such as FSCQ proof generation~\citep{qin2025llmverif} target proof completion in verified system software.
All these benchmarks provide reference implementations and evaluate only the agent's ability to discharge proof obligations.
This setting omits a fundamental challenge of repository-scale verified code generation, since the choice of implementation directly affects which proofs are tractable and the two artifacts must be coherent across the entire repository.
\benchmark{} addresses this by supporting both proof-only and code-and-proof modes, with the latter requiring agents to jointly synthesize implementations and machine-checked proofs across the repository, exposing the implementation-proof coupling that prior repository-scale benchmarks abstract away.
To our knowledge, \benchmark{} is the first benchmark to evaluate joint code-and-proof generation at the repository level.

\smallsec{Third, prior benchmarks are limited in scope and face contamination risks.}
A large body of Lean~4 benchmarks targets formalized mathematics rather than software, including LeanDojo~\citep{yang2023leandojo}, LeanAgent~\citep{kumarappan2024leanagent}, and APE-Bench~\citep{xin2025ape}, all built on the Mathlib library~\citep{mathlib}.
Although these benchmarks have driven impressive progress in neural theorem proving, the structure of mathematical proof differs substantially from the verification of imperative software, where reasoning is coupled to executable definitions, data types, and effectful behavior.
A separate concern is the contamination of training data, since reference solutions for many widely used program sources are freely available online and may have been included in the pretraining data of evaluated models~\citep{jain2024livecodebench,riddell2024quantifying}.
Benchmarks that directly reuse such sources inherit this exposure when the underlying problems and their reference solutions remain widely accessible.
\benchmark{} addresses this by sourcing tasks from real-world repositories across multiple source languages and producing novel Lean~4 formalizations through manual curation, so that no Lean~4 ground-truth proof for any benchmark instance is directly available.
This translation from source languages to Lean~4 is itself novel curation work, providing a structural contamination guarantee that complements the diversity of domains and difficulty levels in the corpus.

\smallsec{Finally, prior benchmarks rarely consider evaluation protocols designed for agent-based settings.}
Most existing benchmarks evaluate raw LLM outputs in single-shot or simple iterative settings, without the tool access that deployed coding agents use in practice.
In contrast, \benchmark{} evaluates frontier coding agents with full tool access.
Furthermore, prior benchmarks provide no mechanism to detect or correct errors in their own ground truth, a concern recently surfaced in formal verification benchmarks~\citep{feng2026certified}.
When a specification is wrong or a reference implementation is incorrect, it becomes impossible to distinguish agent failures from benchmark defects.
Prior repository-scale benchmarks also lack anti-cheating measures, leaving open whether reported results reflect genuine verification capability or reward hacking.
\benchmark{} addresses these through a formal audit mechanism that accepts machine-checked negative evidence (\Cref{sec:benchmark:audit}) and anti-cheating safeguards that prevent reward hacking through axiom injection or definition edits (\Cref{sec:benchmark:anticheat}).

\section{The \benchmark Benchmark}
\label{sec:benchmark}

\begin{figure}[ht]
\centering
\includegraphics[width=\linewidth]{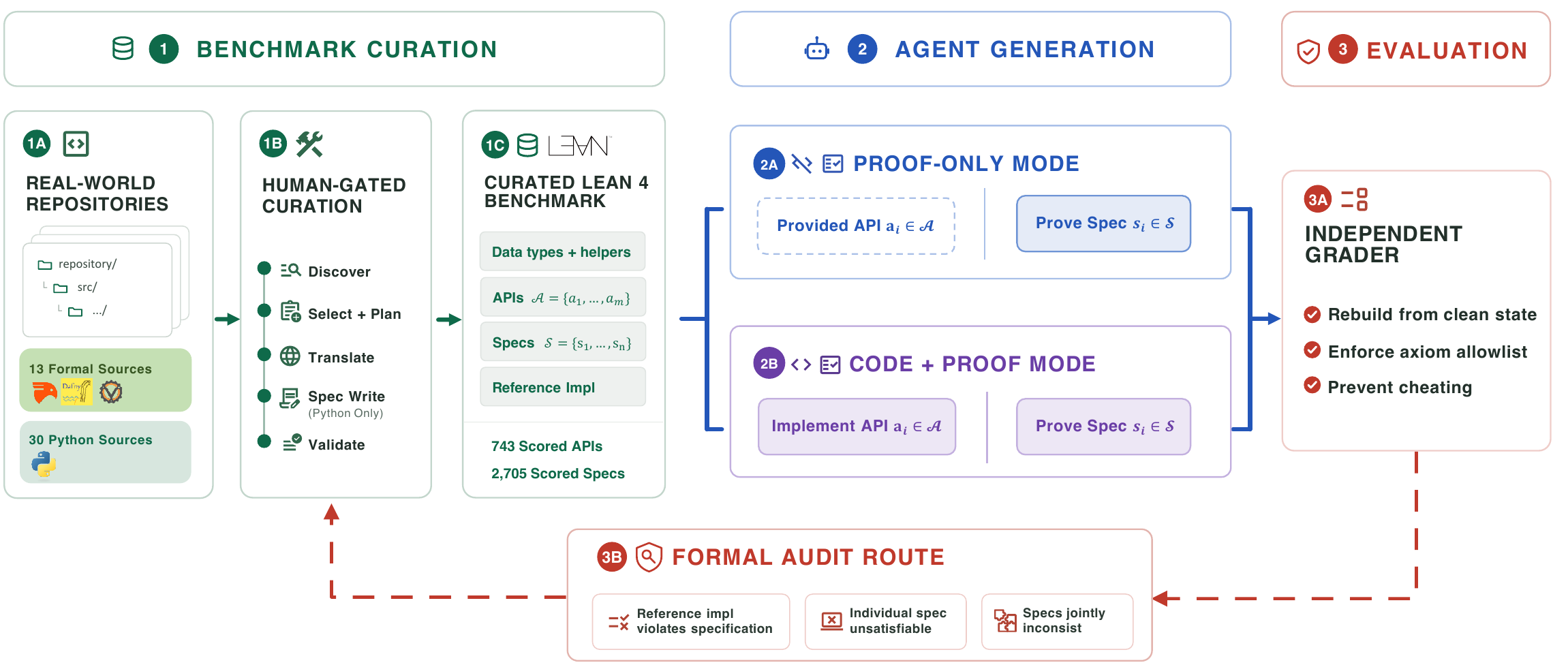}
\caption{\textbf{\benchmark{}'s end-to-end construction and evaluation workflow.} Human-gated curation converts real-world Python and formal-language repositories into Lean 4 benchmark instances with fixed definitions, API signatures, specifications, and reference implementations. Agents are evaluated in proof-only or code-and-proof mode and scored by an independent grader, while the formal audit route returns machine-checked evidence of benchmark defects to curation.}
\label{fig:pipeline}
\end{figure}

\Cref{fig:pipeline} summarizes \benchmark{} from benchmark curation to agent evaluation.
We first define the instance format (\Cref{sec:benchmark:dataformat}) and two task modes (\Cref{sec:benchmark:eval}), then describe the curation pipeline (\Cref{sec:benchmark:curation}) and resulting dataset (\Cref{sec:benchmark:dataset}), and finally introduce the formal audit mechanism (\Cref{sec:benchmark:audit}).






\begin{figure}[t]
\begin{lstlisting}[language=lean]
-- API signatures
abbrev CreateAccountSig := AccountId -> Ledger -> Ledger
abbrev AccountExistsSig := AccountId -> Ledger -> Bool
abbrev GetBalanceSig    := AccountId -> Ledger -> Option Balance

-- Interface structure: one field per API signature
structure RepoImpl where
  createAccount : CreateAccountSig
  accountExists : AccountExistsSig
  getBalance    : GetBalanceSig

-- Canonical implementation (instantiated to reference implementation in proof-only mode and agent solution in code-and-proof mode)
def canonical : RepoImpl := {
  createAccount := Bank.createAccount,
  accountExists := Bank.accountExists,
  getBalance    := Bank.getBalance,
}

-- Specification: a predicate over implementations
def spec_create_zero_balance (impl : RepoImpl) : Prop :=
  ∀ (id : AccountId) (ledger : Ledger),
    impl.accountExists id ledger = false ->
    impl.getBalance id (impl.createAccount id ledger) = some 0

-- Theorem: Proof obligation on the canonical implementation
theorem proof_create_zero_balance : spec_create_zero_balance canonical := by sorry
\end{lstlisting}
\caption{\textbf{Excerpt from a \benchmark{} example illustrating a benchmark instance.} Frozen content like API signatures and specifications are provided by the curator. The agent fills in the body of \texttt{canonical} (in code-and-proof mode) and the proof obligation marked \texttt{sorry} (in both mode).}
\label{fig:format-snippet}
\end{figure}

\subsection{Data Format}
\label{sec:benchmark:dataformat}

A \benchmark{} instance is a multi-module Lean~4 project where the curator provides three layers of fixed content, namely data type and helper definitions, API signatures, and formal specifications.
The agent's task is to discharge two kinds of obligations within this scaffold, an \emph{implementation obligation} for each API signature and a \emph{proof obligation} for each specification.
Formally, a \benchmark{} instance contains following components:
\begin{enumerate}[leftmargin=*]
  \item A collection of \textbf{data type and helper definitions} shared across the rest of the instance.
  \item A set of \textbf{API signatures} $\mathcal{A} = \{a_1, \dots, a_m\}$, each with a reference implementation.
  \item An \textbf{interface structure type} $\repoimpl$ that collects all required API implementations, with one field of type $a_i$ for each $a_i \in \mathcal{A}$.
  \item A set of \textbf{specifications} $\mathcal{S} = \{S_1, \dots, S_n\}$, each of type $\repoimpl \to \mathtt{Prop}$.
  \item A \textbf{canonical implementation} $\canonical : \repoimpl$, populated by the reference implementations in \modeproof
  mode and filled in by the agent in \modecodeproof mode.
\end{enumerate}

Crucially, each specification is parameterized over $\repoimpl{}$ instead of fixed to a single implementation.
This design allows flexible switch of the implementation target to prove, and is the key to support the audit mechanism explained in~\Cref{sec:benchmark:audit}.
\Cref{fig:format-snippet} illustrates the format on a small example drawn from the a reference instance.

\subsection{Task and Evaluation}
\label{sec:benchmark:eval}

\benchmark{} supports two task modes that differ in which obligations the agent must discharge.
In both modes, we evaluate whether the agent can produce artifacts that prove all specifications for a given \benchmark{} instance.
Full coverage is necessary because any unproven specification leaves room for a bug it would have caught, and because specifications vary widely in difficulty, so partial coverage can be inflated by easy ones.

\paragraph{Proof-only mode.}
In this mode, we measure whether the agent can prove every specification against the reference implementation, and $\canonical$ is instantiated from reference implementation.
Formally, for each specification $S_i \in \mathcal{S}$, the agent must produce a machine-checked Lean~4 proof of $S_i(\canonical)$.

\paragraph{Code-and-proof mode.}
In this mode, we measure whether the agent can both implement every API and prove every specification against its own implementations.
$\canonical$ in this mode is instantiated from the agent's own implementations.
Formally, the agent must both provide a function body of type $a_i$ for each $a_i \in \mathcal{A}$ and produce a machine-checked Lean~4 proof of $S_i(\canonical)$ for each $S_i \in \mathcal{S}$.

\paragraph{Code-and-proof is qualitatively different.}
\benchmark{} is, to our knowledge, the first repository-scale benchmark to require joint implementation and proof synthesis.
Existing repository-scale benchmarks fix the implementation and evaluate proof generation in isolation, ignoring how implementation and proof choices interact in real verification work.
The agent can refactor implementations toward proof-friendly forms, but implementation changes can also invalidate proofs across modules.
The two modes therefore measure qualitatively different yet complementary capabilities of formal verification.

\paragraph{Preventing reward hacking.} \label{sec:benchmark:anticheat}

To ensure the benchmark evaluates genuine verification work, the grader prevents both edits to benchmark content and proof-trivializing mechanisms.
First, we use explicit markers to constrain the regions agents may modify.
The grader extracts only contents from these regions and inserts them into a fresh copy of the source benchmark before grading.
Second, an axiom allowlist rejects proofs that depend on agent-introduced or untrusted axioms.
Finally, a rule-based detector and an LLM judge screen agent-introduced declarations for mechanisms that can trivialize proofs or decouple logical semantics from executable behavior, including malicious typeclass instances and noncomputable choice combined with \texttt{@[implemented\_by]} (e.g., through \texttt{Exists.choose}).
We discuss these anti-cheating mechanisms in Appendix~\ref{app:anticheat}.

\subsection{\benchmark{} Curation Pipeline}
\label{sec:benchmark:curation}

To construct \benchmark{}, we developed a multi-stage pipeline that converts source repositories into \benchmark{} instances.
Each stage runs as an LLM agent and is reviewed by a human curator before the next stage begins.
The pipeline supports two source tracks, distinguished by the kind of curation work each repository requires.
\emph{Track~1} covers repositories already written in a formal language such as Dafny, Verus, or Coq, where we need to translate existing types, definitions, and specifications into Lean~4.
\emph{Track~2} covers non-formal repositories, typically Python, where we translate each implementation into Lean~4 and also write the specifications based on original repositories and documentations.
The curation pipeline and human reviewers ensure that the resulting Lean projects form well-defined benchmark instances while remaining faithful to the structure and intended behavior of the source repositories.

\paragraph{Repository selection.}
We select repositories using four criteria: real-world provenance, domain diversity, a wide range of scale and difficulty, and feasibility of faithful translation into a Lean scaffold of moderate size.
Track~1 draws from actively used verification projects, while Track~2 draws from high-star, widely used Python libraries, together spanning smart contracts and blockchain protocols, distributed systems, security-critical parsing and encoding, formalized mathematical algorithms, and core data structures and algorithms.

\paragraph{Pipeline stages.}
The pipeline runs the following stages, and each stage requires curator approval before proceeding.
The first three stages, \emph{discover}, \emph{select}, and \emph{plan}, list all available declarations from the source repository, identify which ones to include, compute their dependencies, and commit to specific Lean signatures and a module layout.
The \emph{translate} stage orchestrates per-module translations, dispatching each module to an executor agent in dependency order to enable parallel and scoped translation.
The curator reviews each translated definition, and items that fail review are re-translated.
On Track~2, an additional \emph{spec writing} stage drafts specifications in natural language for curator review and then formalizes the approved set in Lean.
We require specifications in Track~2 are grounded in the upstream documented behavior and test suites and manually reviewed for semantic fidelity, and ensure that each API is constrained by multiple interacting specifications.
The final \emph{validate} stage runs automated structure and build checks alongside LLM-judged semantic reviews of specification intent, idiomatic Lean usage, and other aspects.

\paragraph{Reusability.} \label{sec:benchmark:reuse}
We build the curation pipeline stages by creating modular agent skills~\citep{anthropic2026skills}, including source-language translation rules, per-stage agent behaviors, and validation prompts.
We currently provide source-language skills for Python, Dafny, Verus, and Coq.
Because the pipeline stages, validation logic, and Lean output format are shared across tracks, adding a new source language requires only a new skill.
This makes \benchmark{} directly extensible to more source languages and verification domains.


\subsection{Dataset Overview}
\label{sec:benchmark:dataset}

\begin{table}[ht]
\centering
\caption{\textbf{Summary statistics of \benchmark{} by source track.} We report the per-instance mean and maximum for each metric. Source LoC is the lines of valid code of upstream repositories selected for curation, excluding blank lines and comments.}
\label{tab:corpus-overview}
\small
\setlength{\tabcolsep}{4pt}
\begin{tabular}{llr cc cc cc}
\toprule
& & & \multicolumn{2}{c}{APIs} & \multicolumn{2}{c}{Specs} & \multicolumn{2}{c}{Source LoC} \\
\cmidrule(lr){4-5} \cmidrule(lr){6-7} \cmidrule(lr){8-9}
Track & Source languages & Inst. & mean & max & mean & max & mean & max \\
\midrule
Track~1 (formal)     & Dafny, Verus, Coq & 13 & 36.0 & 88 & 92.8 & 203 & 7{,}759 & 56{,}887 \\
Track~2 (non-formal) & Python            & 30 &  9.2 & 71 & 50.0 & 109 &   793 &  4{,}047 \\
\midrule
\textbf{Overall}     &                   & \textbf{43} & \textbf{17.3} & \textbf{88} & \textbf{62.9} & \textbf{203} & \textbf{2{,}899} & \textbf{56{,}887} \\
\bottomrule
\end{tabular}
\end{table}

\benchmark{} consists of 43 instances curated from real-world repositories, with 13 from Track~1 (formal) and 30 from Track~2 (Python).
The evaluated corpus totals 743 scored APIs and 2{,}705 scored specifications.
\Cref{tab:corpus-overview} reports per-track summary statistics.

The instances span a broad range of verification-relevant code, including smart contracts and blockchain protocols, distributed systems and consensus, security-critical infrastructure, formal-mathematics libraries, data structures, algorithm collections, and numerical utilities.
They also vary widely in scale, from a few hundred to over fifty thousand source lines per repository and from single digits to over 100 specifications per instance, reflecting a corresponding range of verification difficulty.
This diversity across domains and difficulty levels supports comprehensive assessment of an agent's formal-verification capability.

\paragraph{Contamination.}
A central concern for benchmarks built on public code is contamination, the risk that an evaluated model has memorized the reference solutions during pretraining.
\benchmark{} is largely free of this concern by construction.
The Lean~4 specifications, implementations, and proofs are all novel curation work with no public Lean~4 ground truth.
Track~1 upstream content exists only in other formal languages, and Track~2 upstream content is Python without specifications or proofs.

\subsection{Benchmark Formal Audit Mechanism}
\label{sec:benchmark:audit}

Even carefully curated formal-verification benchmarks can contain latent errors, including specifications that no implementation satisfies, reference implementations that fail their intended specifications, and specification sets that are mutually inconsistent.
Such errors are subtle and can survive type-checking, builds, and curator review, but they can surface when an agent attempts to prove.
Treating these silent failures as agent errors conflates benchmark defects with agent capability.

\benchmark{} addresses this with an \emph{audit mechanism}, a separate formal-evidence path that accepts three forms of machine-checked negative evidence:
\begin{align}
& \neg \textstyle\bigwedge_{S \in \mathcal{S}}S(\canonical), \label{eq:disprove} \\
& \neg \exists\,\textit{impl} : \repoimpl,\ S(\textit{impl}) \text{ for some } S \in \mathcal{S}, \label{eq:unsat}\\
& \forall S \in \mathcal{S}', \exists\,\textit{impl} : \repoimpl,\ S(\textit{impl}) \text{ with } \neg \exists\,\textit{impl},\ \textstyle\bigwedge_{S \in \mathcal{S}'} S(\textit{impl}) \text{ for some } \mathcal{S}' \subseteq \mathcal{S}. \label{eq:joint_unsat}
\end{align}
\Cref{eq:disprove} catches reference implementation bugs in \modeproof mode.
In \modecodeproof mode, \Cref{eq:unsat} catches individually unsatisfiable specifications $S$, and~\Cref{eq:joint_unsat} catches inconsistent specification sets while preserving evidence that each member remains individually satisfiable.

During benchmark curation, we used the audit mechanism alongside evaluated agents to iteratively improve benchmark quality.
We present details case studies on how this mechanism improves the benchmark quality in~\Cref{app:audit}.
The audit mechanism is also valuable in supporting continuous benchmark improvement as future agents grow more capable.


\section{Evaluation}
\label{sec:eval}

In this section, we evaluate four frontier coding-agent configurations on \benchmark{} under both task modes and analyze where current agents succeed and where they fall short.

\subsection{Experimental Setup}
\label{sec:eval:setup}

We evaluate two coding-agent harnesses with frontier models.
Codex (v0.140.0) is paired with two GPT-5.5 profiles: the default reasoning effort (\texttt{medium}) and the \texttt{xhigh} reasoning effort.
Claude Code (v2.1.191) is paired with Claude Opus~4.8 and Claude Sonnet~5 at \texttt{xhigh} reasoning effort.
All agents run with full tool access, including file-system edits, build invocations, and the Lean toolchain.
We report the count of fully solved instances in both \modecodeproof and \modeproof modes.
We adopt full solve rather than partial specification coverage because partial coverage can be inflated by easy specifications, and in \modecodeproof mode an unproven specification may indicate not merely a missing proof but an implementation that violates it, so only full coverage certifies the agent's code as correct.
We use Lean v4.29.1 throughout the evaluation.
The detailed setup is described in~\Cref{app:setup}.

\subsection{Evaluation Results and Analysis}

\begin{figure}[tph]
\centering
\includegraphics[width=\textwidth]{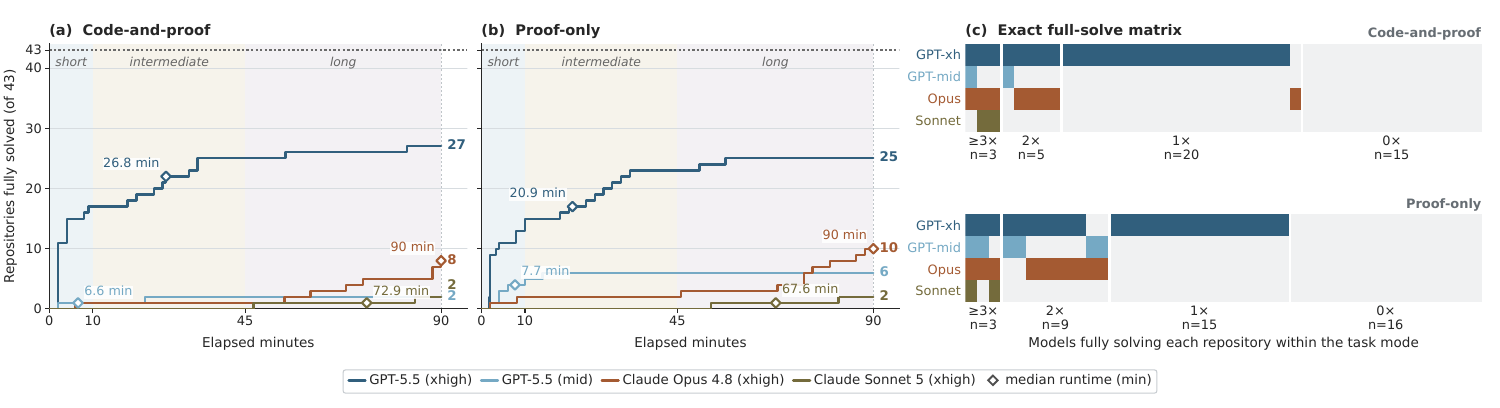}
\caption{\textbf{Agent performance on \benchmark{}.}
(a,b) Cumulative full solves, out of 43, over the 90-minute budget in \modecodeproof and \modeproof; diamonds mark each configuration's median runtime.
(c) Which configurations fully solve each instance in each mode.}
\label{fig:eval-overview}
\end{figure}

\paragraph{GPT-5.5 (xhigh) leads both modes, but \benchmark{} remains frontier-resistant.}
\label{sec:eval:overall}
GPT-5.5 (xhigh) fully solves 27 of 43 instances in \modecodeproof and 25 in \modeproof, ahead of Claude Opus~4.8 (8 and 10), GPT-5.5 (mid) (2 and 6), and Claude Sonnet~5 (2 in each mode) (\Cref{fig:eval-overview}(a,b)).
It also progresses fastest, reaching 25 and 23 full solves within 45 minutes.
Nevertheless, 10 instances resist all eight configurations across both modes, and most solved instances are closed by only one configuration (\Cref{fig:eval-overview}(c)).
The bottleneck is not local proof skill: GPT-5.5 (xhigh) passes 87.3\% of specifications in \modecodeproof and 85.8\% in \modeproof.
High per-specification coverage therefore does not imply repository completion.
The specifications left are mostly hardest to prove, where agents must discover shared invariants, organize them into a reusable lemma library, and preserve that library in a buildable artifact.

\begin{figure*}[h]
\centering
\includegraphics[width=\textwidth]{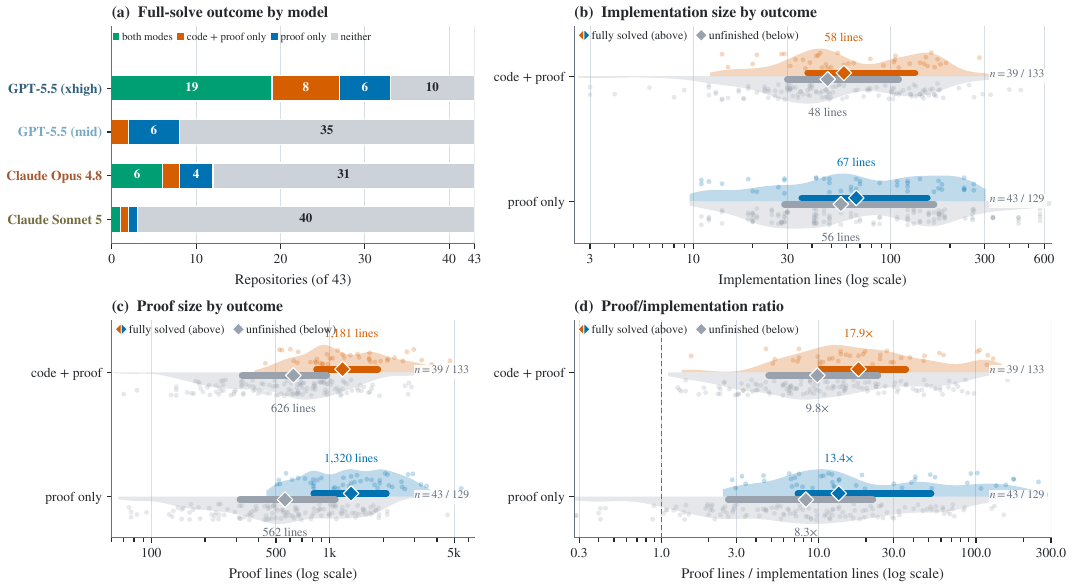}
\caption{\textbf{Full-repository outcomes and artifact sizes by task mode.}
(a) Paired full-solve outcomes across the two modes for each of the 172 instance--agent pairs.
(b--d) Implementation lines, proof lines, and their ratio at the end of each of the 344 runs, split by mode and by whether the run is a full solve.
Diamonds mark medians and thick segments mark interquartile ranges.}
\label{fig:eval-modes}
\end{figure*}

\paragraph{Implementation freedom matters through the proof target; proof volume is what separates full solves.}

Across the 172 matched instance--agent pairs, 26 are fully solved in both modes, 13 only in \modecodeproof, 17 only in \modeproof, and 116 in neither (\Cref{fig:eval-modes}(a)).
We manually examined agent solutions in \modecodeproof, and found five instance-agent pairs across three repositories where the agent replaces a hard-to-verify reference algorithm with a simpler implementation of its own that satisfies the same specifications~(\Cref{app:case:implementation-substitution}).
Each substitution is behaviorally correct rather than a shortcut, so what the agent gains is provability and what it gives up is efficiency.
Across these five pairs the agent closes all 250 specifications, whereas proving against the fixed reference in \modeproof closes only 201.
Conversely, we find 17 pairs that are full solves in \modeproof but not in \modecodeproof, where we see implementation adds difficulty for models.
The size statistics in \Cref{fig:eval-modes}(b--d) show that fully solved and unfinished runs write similar amounts of implementation code, while full solves contain roughly twice the proof text and correspondingly higher proof-to-implementation ratios.
Completing a repository-level formal verification is a matter of sustained proof work rather than implementation volume.

\begin{figure*}[tph]
\centering
\includegraphics[width=\textwidth]{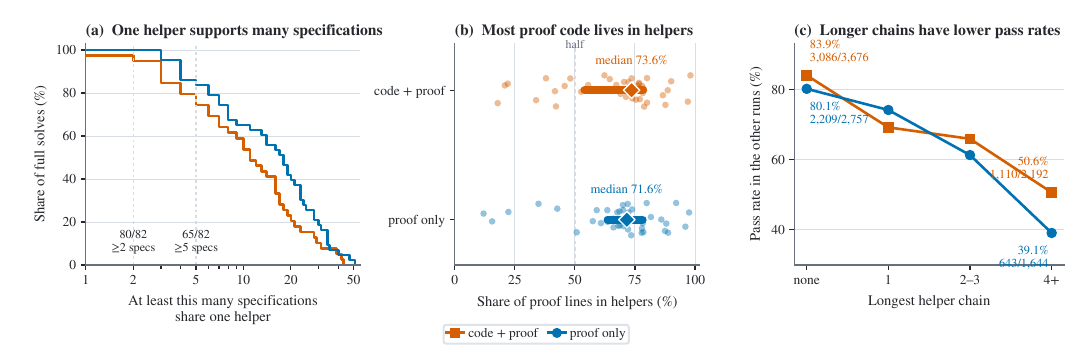}
\caption{\textbf{Proof structure in the 82 full solves.}
A helper is a theorem the agent writes to support its specification proofs.
(a) How many specifications share one helper.
(b) Share of proof lines that live in helpers.
(c) Pass rates in the other seven runs, grouped by each specification's helper-chain depth in GPT-5.5 (xhigh)'s proof-only solutions.}
\label{fig:eval-proof-architecture}
\end{figure*}

\paragraph{Completed repositories rely on shared lemma libraries.}
Across the 82 full solves, agent-written helper theorems contain a median of 73.6\% of proof lines in \modecodeproof and 71.6\% in \modeproof (\Cref{fig:eval-proof-architecture}(b)).
These lemmas are not for individual proof goal but serve as shared libraries, as 80 of the 82 solves share a helper across at least two specifications and 65 across at least five (\Cref{fig:eval-proof-architecture}(a)).
Completed repositories are organized around reusable lemma libraries rather than independent proofs per specification.

\paragraph{Deep lemma chains predict where other agents fail.}
We measure each specification's helper-chain depth in one GPT-5.5 (xhigh) \modeproof full solve and check how the same specifications fare in the other seven runs (\Cref{fig:eval-proof-architecture}(c)).
Specifications needing no helper pass at 83.9\% in \modecodeproof and 80.1\% in \modeproof, falling to 50.6\% and 39.1\% at depth four or more.
Deep specifications are hard because they require layered and global reasoning, where the agent must first build a chain of prerequisite lemmas about lower-level definitions.

\begin{figure*}[h]
\centering
\includegraphics[width=\textwidth]{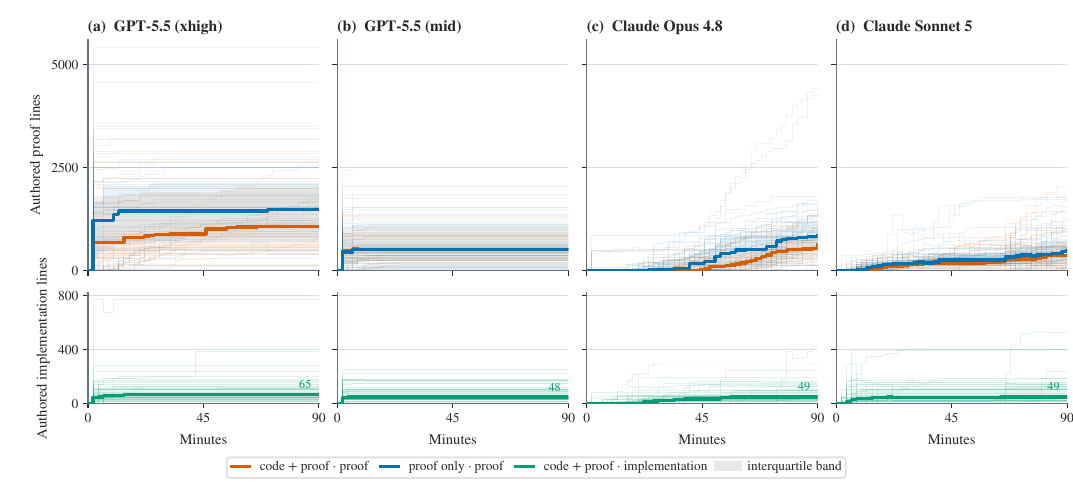}
\caption{\textbf{Agents fix implementations early and grow proofs until the deadline.}
Columns are agent configurations.
Top row shows authored proof lines over time in both modes; bottom row shows authored implementation lines in \modecodeproof.
Thin lines are per-repository trajectories, bold lines are medians, and bands are interquartile ranges.
Counts exclude blank lines, comments, and placeholders.}
\label{fig:eval-dynamics}
\end{figure*}

\begin{figure*}[tph]
\centering
\includegraphics[width=\textwidth]{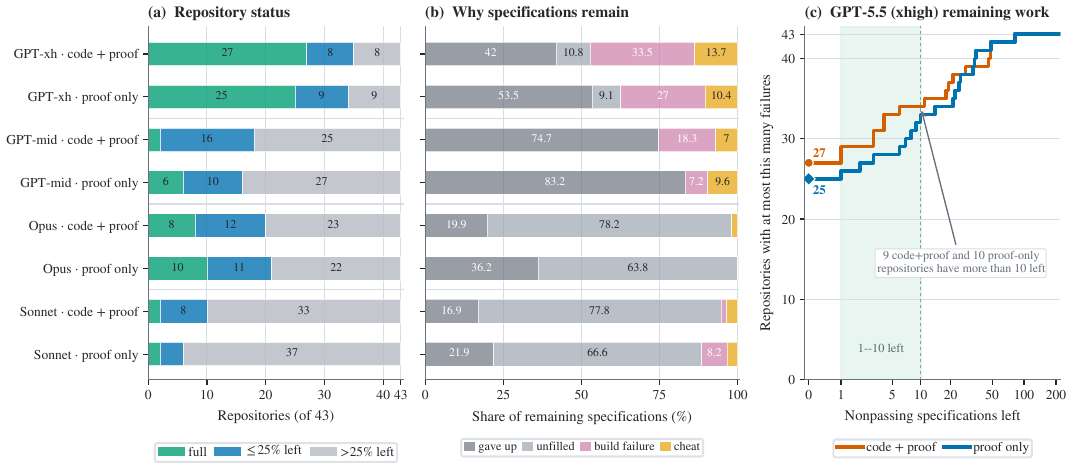}
\caption{\textbf{Where failures remain at the end of a run.}
(a) Repositories grouped by the share of their specifications still failing.
(b) Breakdown of remaining specifications by failure reason.
(c) Number of GPT-5.5 (xhigh) repositories with at most $x$ failing specifications.}
\label{fig:eval-failures}
\end{figure*}

\paragraph{Agents commit to an implementation early and spend the rest of the budget on proofs.}
Every configuration reaches its final median implementation size within the first half of the run, while proof text keeps growing until the deadline (\Cref{fig:eval-dynamics}).
GPT-5.5 (xhigh) fixes its median 65 implementation lines by minute 30 and then grows proof text from 883 to 1{,}077 lines, and Claude Sonnet~5 adds only 5 implementation lines after minute 60 while its proof text nearly doubles.
The agents differ mainly in pace, as GPT-5.5 (xhigh) plateaus early while Claude Opus~4.8 and Sonnet~5 are still writing proofs when time expires, suggesting their lower solve counts partly reflect the time budget.
The one-way ordering also shows that agents treat the implementation as fixed scaffolding rather than a lever, so when proofs get stuck they grind on the proof layer instead of returning to refactor the code into a more provable form.

\paragraph{Strong agents attempt the hardest obligations but cannot close them.}
GPT-5.5 (xhigh) ends with at most a quarter of each repository's specifications failing in 35 of 43 \modecodeproof repositories and 34 of 43 in \modeproof, more than any other configuration (\Cref{fig:eval-failures}(a)).
The failure reasons also differ by strength: a third of its remaining specifications fail at build time and a further 14\% are rejected as cheating, whereas Claude Opus~4.8 and Claude Sonnet~5 leave roughly 78\% of theirs with no proof body at all (\Cref{fig:eval-failures}(b)).
Strong agents thus fail while attempting the hardest obligations, whereas weaker agents never reach them.
The remaining failures are not near misses, since only 6 of its 16 unfinished \modecodeproof repositories are within five specifications of completion and nine have more than ten left (\Cref{fig:eval-failures}(c)).
The residual specifications typically form a cluster that shares a few missing invariants, so closing them requires the inductive generalization the agent never found, not additional proof attempts.

\begin{figure*}[tph]
\centering
\includegraphics[width=\textwidth]{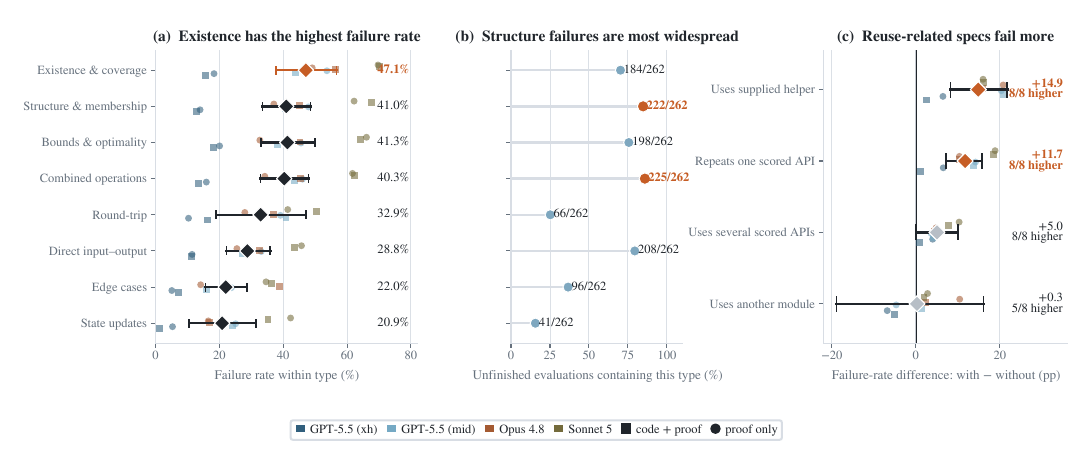}
\caption{\textbf{Which specifications fail.}
(a) Failure rate by semantic type, equalized across repositories.
(b) How often each type appears among the 262 unfinished evaluations.
(c) Within-repository failure-rate differences for four statement features.
Diamonds and whiskers show aggregate estimates with bootstrap 95\% confidence intervals; small markers show the eight agent and mode configurations.
Types and features may overlap.}
\label{fig:eval-spec-taxonomy}
\end{figure*}

\paragraph{Specifications about global properties and definition reuse fail most.}
Existence-and-coverage specifications, which assert that an output covers or accounts for all relevant inputs, have the highest failure rate at 47.1\% and stay highest under every sensitivity analysis (\Cref{fig:eval-spec-taxonomy}(a)).
These are global properties that cannot be closed by case-splitting on a single call, matching the inductive-reasoning gap identified above.
Within repositories, specifications that use a supplied helper definition or call one API repeatedly fail 14.9 and 11.7 points more often, consistently across all eight configurations, while merely referencing several APIs or another module adds little (\Cref{fig:eval-spec-taxonomy}(c)).
Difficulty therefore comes from unfolding and reasoning about supplied definitions and iterated behavior, not from crossing module boundaries.

\section{Conclusion}
\label{sec:conclusion}
We introduced \benchmark{}, the first benchmark for joint implementation and proof synthesis at the repository level in Lean~4.
\benchmark{} comprises 43 multi-module Lean~4 instances curated from real-world repositories in Dafny, Verus, Coq, and Python, together with an extensible curation pipeline, an evaluation harness for both task modes, and a formal audit mechanism that turns machine-checked negative evidence into benchmark corrections.
Our evaluation shows that \benchmark{} is frontier-resistant, as the strongest configuration fully solves only 27 of 43 instances in \modecodeproof mode and 10 instances resist every configuration in both modes.
The gap is not local proof skill but repository-scale organization.
The strongest agent pass over 80\% of individual specifications yet fails to discover shared invariants, build the reusable lemma libraries that deep specifications require, and keep the whole repository consistent and buildable.
We hope \benchmark{} serves as a concrete testbed for driving agents toward producing fully verified software at practical scale.

\paragraph{Limitations.}
We discuss the following limitations.
First, \benchmark{} currently targets Lean~4 only.
We chose Lean as the initial target because it is a particularly active ecosystem for LLM-based formal reasoning, providing a favorable setting for current agents.
At the same time, Track~1 translates projects from Dafny, Verus, and Coq, so their specification and invariant structures are represented in the benchmark, and the curation pipeline is extensible to other target languages (\Cref{sec:benchmark:reuse}).
Second, the corpus favors code that translates cleanly into Lean, and the main absent class is concurrent or temporal protocols whose upstream formalizations do not port to a Lean scaffold of moderate size.
Benchmarking formal verification of incremental maintenance tasks is also a valuable future direction.
Finally, the audit mechanism certifies formal satisfiability but cannot ensure that specifications are semantically correct or complete, so all specifications are manually reviewed and Track~1 specifications are additionally cross-checked against the source-language formalizations.

\begin{ack}
This work is partially supported by a grant from Amazon.
Any opinions, findings, and conclusions or recommendations expressed in this material are those of the authors and do not necessarily reflect the views of Amazon.
\end{ack}

\bibliographystyle{plain}
\bibliography{references}

\clearpage

\appendix

\input{appendix}



\end{document}

%% file: appendix.tex
\section{Experimental Configuration}
\label{app:setup}

\subsection{Agent Harnesses and Models}

Codex (v0.140.0) is invoked with GPT-5.5.
The reasoning-effort setting is the Codex-side parameter \texttt{model\_reasoning\_effort}; the default value (\texttt{medium}) corresponds to the GPT-5.5~(mid) row in~\Cref{sec:eval} and the \texttt{xhigh} value corresponds to the GPT-5.5~(xhigh) row.
Claude Code (v2.1.191) is invoked with Claude Opus~4.8 and Claude Sonnet~5, both at \texttt{xhigh} reasoning effort.
All agents have access to the project's Lean toolchain in version v4.29.1 and to \texttt{lake build} commands.

\subsection{Evaluation Cost}
 
\Cref{tab:cost} reports the aggregate evaluation cost for each (agent, mode) cell, summed over all $43$ instances.
 
\begin{table}[ht]
\centering
\caption{Aggregate evaluation cost (USD) per cell, summed over all $43$ instances.}
\label{tab:cost}
\small
\begin{tabular}{l r r}
\toprule
Agent & Code-and-proof & Proof-only \\
\midrule
GPT-5.5 (xhigh)  & \$2{,}865 & \$2{,}964 \\
GPT-5.5 (medium) & \$928     & \$1{,}094 \\
Claude Opus 4.8  & \$1{,}983 & \$2{,}279 \\
Claude Sonnet 5  & \$633     & \$791 \\
\bottomrule
\end{tabular}
\end{table}

\clearpage

\section{Benchmark Sources and Licenses}
\label{app:sources}
 
\Cref{tab:sources} lists the upstream source, language, and license for each of the 43 active instances.
All Track~1 instances are translated from formal-source projects (Dafny, Verus, Coq); all Track~2 instances are translated from Python.
Each instance's manifest records the upstream repository, license, and pinned commit hash.
Three active instances derive from copyleft upstreams: \texttt{flocq} (LGPL-3.0-or-later), \texttt{huffman} (LGPL-2.1-or-later), and \texttt{portion} (LGPL-3.0-or-later).
The \texttt{arithmetic} instance is translated from the \texttt{verus-lang/verus} example suite.

\begin{table}[ht]
\centering
\caption{Upstream sources, languages, licenses, and pinned commits for the 43 active instances. Commit hashes are short SHAs of the upstream snapshot used by the curation pipeline. Every instance is translated into a novel Lean~4 formalization; no Lean~4 ground truth is publicly available.}
\label{tab:sources}
\resizebox{\textwidth}{!}{
\begin{tabular}{l l l l l}
\toprule
Instance & Upstream source & Lang. & License & Commit \\
\midrule
\multicolumn{5}{l}{\emph{Track 1: Formal-source instances}} \\
\texttt{arithmetic}          & \href{https://github.com/verus-lang/verus}{verus-lang/verus} (vstd)              & Verus & MIT & \texttt{8c06fbd7} \\
\texttt{dedekind\_reals}     & \href{https://github.com/rocq-community/dedekind-reals}{rocq-community/dedekind-reals} & Coq & MIT & \texttt{da4a7452} \\
\texttt{deposit\_sc}         & \href{https://github.com/ConsenSys/deposit-sc-dafny}{ConsenSys/deposit-sc-dafny}    & Dafny & Apache-2.0 & \texttt{cf321d10} \\
\texttt{flocq}               & \href{https://gitlab.inria.fr/flocq/flocq}{gitlab.inria.fr/flocq/flocq}            & Coq   & LGPL-3.0-or-later & \texttt{7aab8f55} \\
\texttt{huffman}             & \href{https://github.com/rocq-community/huffman}{rocq-community/huffman}           & Coq   & LGPL-2.1-or-later & \texttt{cc7d4cc4} \\
\texttt{json}                & \href{https://github.com/dafny-lang/libraries}{dafny-lang/libraries}               & Dafny & MIT & \texttt{b486ff7f} \\
\texttt{piggybank}           & \href{https://github.com/AU-COBRA/ConCert}{AU-COBRA/ConCert}                       & Coq   & MIT & \texttt{341f440a} \\
\texttt{sequences}           & \href{https://github.com/dafny-lang/libraries}{dafny-lang/libraries}               & Dafny & MIT & \texttt{b486ff7f} \\
\texttt{unicode}             & \href{https://github.com/dafny-lang/libraries}{dafny-lang/libraries}               & Dafny & MIT & \texttt{b486ff7f} \\
\texttt{verdict}             & \href{https://github.com/secure-foundations/verdict}{secure-foundations/verdict}    & Verus & MIT / Apache-2.0 & \texttt{9bc18bc5} \\
\texttt{verified\_bitmasks}  & \href{https://github.com/achreto/verified-bitmasks}{achreto/verified-bitmasks}     & Verus & MIT & \texttt{cce6985c} \\
\texttt{verified\_ironkv}    & \href{https://github.com/verus-lang/verified-ironkv}{verus-lang/verified-ironkv}   & Verus & MIT & \texttt{08be20c3} \\
\texttt{vest}                & \href{https://github.com/secure-foundations/vest}{secure-foundations/vest}          & Verus & MIT & \texttt{db63c23b} \\
\midrule
\multicolumn{5}{l}{\emph{Track 2: Python-source instances}} \\
\texttt{base58}              & \href{https://github.com/keis/base58}{keis/base58}                                 & Python & MIT & \texttt{2fae7065} \\
\texttt{cachetools}          & \href{https://github.com/tkem/cachetools}{tkem/cachetools}                         & Python & MIT & \texttt{48284d73} \\
\texttt{croniter}            & \href{https://github.com/kiorky/croniter}{kiorky/croniter}                         & Python & MIT & \texttt{9810279c} \\
\texttt{difflib}             & \href{https://github.com/python/cpython}{python/cpython}                           & Python & PSF-2.0 & \texttt{669299b6} \\
\texttt{dijkstar}            & \href{https://github.com/wylee/dijkstar}{wylee/dijkstar}                           & Python & MIT & \texttt{aa1237a8} \\
\texttt{ecdsa}               & \href{https://github.com/tlsfuzzer/python-ecdsa}{tlsfuzzer/python-ecdsa}           & Python & MIT & \texttt{bff40c6c} \\
\texttt{galoistools}         & \href{https://github.com/sympy/sympy}{sympy/sympy}                                 & Python & BSD-3-Clause & \texttt{2f9c274d} \\
\texttt{greenery}            & \href{https://github.com/qntm/greenery}{qntm/greenery}                             & Python & MIT & \texttt{588f5e30} \\
\texttt{intervaltree}        & \href{https://github.com/chaimleib/intervaltree}{chaimleib/intervaltree}           & Python & Apache-2.0 & \texttt{1bc406e1} \\
\texttt{ipaddress}           & \href{https://github.com/python/cpython}{python/cpython}                           & Python & PSF-2.0 & \texttt{669299b6} \\
\texttt{jsonpatch}           & \href{https://github.com/stefankoegl/python-json-patch}{stefankoegl/python-json-patch} & Python & BSD-3-Clause & \texttt{0b052032} \\
\texttt{linked\_list}        & \href{https://github.com/TheAlgorithms/Python}{TheAlgorithms/Python}               & Python & MIT & \texttt{7a0fee40} \\
\texttt{munkres}             & \href{https://github.com/bmc/munkres}{bmc/munkres}                                 & Python & Apache-2.0 & \texttt{ac8af9e3} \\
\texttt{netaddr}             & \href{https://github.com/netaddr/netaddr}{netaddr/netaddr}                         & Python & BSD-3-Clause & \texttt{d340feab} \\
\texttt{networkx}            & \href{https://github.com/networkx/networkx}{networkx/networkx}                     & Python & BSD-3-Clause & \texttt{19509286} \\
\texttt{ntheory}             & \href{https://github.com/sympy/sympy}{sympy/sympy}                                 & Python & BSD-3-Clause & \texttt{1a2501b3} \\
\texttt{packaging\_version}  & \href{https://github.com/pypa/packaging}{pypa/packaging}                           & Python & Apache-2.0 / BSD-2 & \texttt{dcac24cc} \\
\texttt{portion}             & \href{https://github.com/AlexandreDecan/portion}{AlexandreDecan/portion}           & Python & LGPL-3.0-or-later & \texttt{b771acfa} \\
\texttt{primefac}            & \href{https://github.com/elliptic-shiho/primefac-fork}{elliptic-shiho/primefac-fork} & Python & MIT & \texttt{28adf4aa} \\
\texttt{primepy}             & \href{https://github.com/janaindrajit/primePy}{janaindrajit/primePy}               & Python & MIT & \texttt{9c98276f} \\
\texttt{prolepticgregorian}  & \href{https://github.com/python/cpython}{python/cpython}                           & Python & PSF-2.0 & \texttt{669299b6} \\
\texttt{pyradix}             & \href{https://github.com/mjschultz/py-radix}{mjschultz/py-radix}                   & Python & ISC & \texttt{b5e4e930} \\
\texttt{pythonconstraint}    & \href{https://github.com/python-constraint/python-constraint}{python-constraint/python-constraint} & Python & BSD-2-Clause & \texttt{f1359ff0} \\
\texttt{reedsolo}            & \href{https://github.com/tomerfiliba-org/reedsolomon}{tomerfiliba-org/reedsolomon} & Python & Unlicense / MIT-0 & \texttt{796639ca} \\
\texttt{rsa}                 & \href{https://github.com/sybrenstuvel/python-rsa}{sybrenstuvel/python-rsa}         & Python & Apache-2.0 & \texttt{42b0e14f} \\
\texttt{semver}              & \href{https://github.com/rbarrois/python-semanticversion}{rbarrois/python-semanticversion} & Python & BSD-2-Clause & \texttt{2cbbee31} \\
\texttt{sortedcontainers}    & \href{https://github.com/grantjenks/python-sortedcontainers}{grantjenks/python-sortedcontainers} & Python & Apache-2.0 & \texttt{3ac35863} \\
\texttt{textdistance}        & \href{https://github.com/life4/textdistance}{life4/textdistance}                   & Python & MIT & \texttt{d6a68d61} \\
\texttt{textwrap}            & \href{https://github.com/python/cpython}{python/cpython}                           & Python & PSF-2.0 & \texttt{669299b6} \\
\texttt{toposort}            & \href{https://gitlab.com/ericvsmith/toposort}{ericvsmith/toposort}                 & Python & Apache-2.0 & \texttt{9fcd0437} \\
\bottomrule
\end{tabular}}
\end{table}

\paragraph{Curation effort.}
Hands-on human effort ranges from several hours for straightforward instances to a few days when translation repairs or specification revisions are required, concentrated on source selection, semantic review of translated definitions, Track~2 specification review, and build validation.
Curation consumes approximately \$60 of model usage per instance at standard API rates, including QA and validation runs in both task modes.


\clearpage
\section{Per-Instance Results}
\label{app:per-instance}
 
\Cref{tab:per-instance} reports per-instance pass counts at each agent's final submitted state for every (agent, mode) cell across the 43 active instances.

\input{figures/tab_per_instance.tex}

\clearpage

\section{Anti-Cheating Mechanisms}
\label{app:anticheat}

The grader must decide whether an agent has verified a repository or merely arranged for the checker to agree with it.
\benchmark{} enforces this through three independent layers, described in turn below.
Each layer is deterministic except the final judge, and each rejects submissions that a naive ``does it compile'' check would accept.

\paragraph{Layer 1: slot-scoped re-rendering.}
An agent works in a sandbox containing the full repository, but only the interiors of \texttt{!benchmark} marker pairs are treated as its submission.
The grader extracts those interiors and overlays them onto a fresh copy of the pristine instance before compiling anything.
Every declaration outside a marker pair therefore reverts to the curated original, which removes an entire family of attacks at once: weakening a specification, redefining a frozen helper, relaxing a type signature, or deleting a proof obligation.

\paragraph{Layer 2: axiom allowlist.}
Compilation alone does not establish that a proof is honest, because Lean admits \texttt{sorry} and user-declared axioms as well-typed terms.
After the build succeeds, the grader emits \texttt{\#print axioms} for every scored theorem and classifies its dependency set.
Only the three standard axioms of Lean's logic are accepted, namely \texttt{Classical.choice}, \texttt{propext}, and \texttt{Quot.sound}.
A curator may extend this set per instance through \texttt{manifest.trusted\_axioms}, which is recorded in the released benchmark.
A theorem that depends on \texttt{sorryAx} grades as \texttt{unfilled}, and one that depends on any other axiom grades as \texttt{axiom\_leaked}.
Neither counts as passed.

This layer is not hypothetical.
Across the corpus $368$ specification outcomes are rejected at this stage, and they concentrate in instances with large finite domains: \texttt{base58} ($67$), \texttt{verified\_bitmasks} ($60$), \texttt{reedsolo} ($53$), and \texttt{sortedcontainers} ($52$).
The dominant pattern is \texttt{native\_decide}, which discharges a decidable goal by evaluating it in the compiler rather than in the kernel and records a dependency on \texttt{Lean.ofReduceBool}.
On \texttt{base58}, for instance, GPT-5.5~(xhigh) closes the Base58 alphabet round-trip this way:

\begin{lstlisting}[language=lean, basicstyle=\ttfamily\footnotesize]
theorem b58Idx_b58Char_fin (d : Fin 58) :
    b58Idx (b58Char d.val) = some d.val := by
  native_decide +revert
\end{lstlisting}

The claim is true and the tactic is legitimate Lean, but it is not a kernel-checked proof and it does not generalize beyond the finite domain it enumerates.
Twenty-eight of that cell's $53$ specifications fail for this reason while $25$ pass, which is why the failure-reason breakdown in~\Cref{fig:eval-failures}(b) separates rejected shortcuts from unattempted obligations.

\paragraph{Layer 3: declaration screening.}
The final layer inspects agent-introduced declarations for constructions that decouple logical semantics from executable behavior.
For typeclass instances, a rule-based prefilter raises two signals, a hollow constant body and a \texttt{priority} override that shadows a standard-library instance, and an LLM judge then classifies each flagged site as \texttt{legit} or \texttt{cheat} from a window of surrounding source.
When both signals fire together the site is classified as a cheat without consulting the judge.

A separate deterministic check covers the case where an agent splits the proof target from the runtime function, described last below.

Each pattern is invisible to the first two layers, because each compiles and none introduces an axiom.
We give one example of each.

\emph{Hollow body on a non-trivial typeclass.}
An ordering instance whose body ignores both arguments makes every comparison reduce to \texttt{True}:

\begin{lstlisting}[language=lean, basicstyle=\ttfamily\footnotesize]
instance : LE Float := <fun _ _ => True>
\end{lstlisting}

Any specification phrased over \texttt{<=} then closes by \texttt{decide} regardless of what the implementation computes.
The signal is that the standard meaning of \texttt{LE} on \texttt{Float} is not constant, so a body that discards its arguments cannot be computing it.

\emph{Priority shadowing.}
The same hollow body becomes more dangerous when it displaces a real instance rather than filling a gap:

\begin{lstlisting}[language=lean, basicstyle=\ttfamily\footnotesize]
instance (priority := 2000) : LT Float := <fun _ _ => True>
\end{lstlisting}

Lean resolves standard-library instances at priority around $1000$, so a declared priority of $2000$ takes precedence over \texttt{Float}'s genuine ordering everywhere in the repository, including inside specifications the agent never touched.
Because this pattern combines two independent signals, the prefilter classifies a hollow body together with a priority override as a cheat without consulting the judge.

\emph{Decidability laundering.}
A \texttt{Decidable} instance may assert its own verdict instead of computing one:

\begin{lstlisting}[language=lean, basicstyle=\ttfamily\footnotesize]
instance : DecidableEq (Foo -> Bar) := fun _ _ => .isTrue (by sorry)
theorem prove_x : spec_x canonical := by decide
\end{lstlisting}

Here the axiom check does catch the \texttt{sorry}, but the variant \texttt{.isTrue (by trivial)} on a proposition that happens to be provable in a trivial instance, or \texttt{:= trivial} supplied directly as a \texttt{Decidable} body, leaves no axiom trace while still making \texttt{decide} report \texttt{isTrue}.

\emph{Splitting the proof target from the runtime function.}
The most consequential pattern does not involve typeclasses at all, so we screen it separately.
An agent can define the scored API as a \texttt{noncomputable} copy of the specification's own witness oracle, and then attach a real algorithm through \texttt{@[implemented\_by]}:

\begin{lstlisting}[language=lean, basicstyle=\ttfamily\footnotesize]
noncomputable def findPathCost : FindPathCostSig :=
  fun g s t => if h : exists c, isMinWalkCost g s t c then h.choose else 0
@[implemented_by realBellmanFord] def findPathCost' := findPathCost
\end{lstlisting}

Every optimality proof about \texttt{findPathCost} is then definitionally true, because the function is the specification's answer set, while the compiled binary still runs a correct shortest-path algorithm.
We encountered this during benchmark development on \texttt{dijkstar}, where the submission produced no output difference from the reference across $20{,}440$ generated graphs and would have passed any differential test.
It is also general: the technique defeats any unique-optimum specification suite no matter how the specifications are strengthened, which is why the defense has to live in the grader.
The grader now rejects \texttt{@[implemented\_by]} in any agent-editable \texttt{Impl/} file outright, voiding the run and grading every specification \texttt{impl\_oracle}, since the attribute has no legitimate use in an implementation slot.
Reference definitions on the proof side that are \texttt{noncomputable} without the attribute, such as \texttt{flocq}'s use of \texttt{Classical.choose}, are unaffected.

\section{Additional Evaluation Results}
\label{app:additional-results}

\subsection{Instances Have Shared Walls, Not Random Residuals}
\label{app:walls}

If the specifications an agent fails were determined mostly by how much budget it happened to spend, different configurations would fail on different specifications.
They do not.
On several instances every configuration that gets close fails on exactly the same set.

\texttt{verdict} is the clearest example and also the corpus's nearest miss: Claude Opus~4.8 in \modecodeproof reaches $112$ of $119$ specifications.
Its seven-specification residual is failed by all eight configurations, so it is exactly the intersection of every configuration's failures.
Four of the seven concern wildcard certificate-name matching (\texttt{spec\_match\_name\_wildcard\_base}, \texttt{\_single\_label}, \texttt{\_multi\_label\_rejected}, and \texttt{spec\_permit\_name\_dot\_prefix}) and three concern projecting parsed certificate fields (\texttt{spec\_cert\_from\_parsed\_sig\_alg\_inner}, \texttt{\_outer}, and \texttt{\_validity}).
An instance that no agent completes is thus $94.1\%$ solved by the best one, and the part that resists is a wall two groups wide rather than a scattering of hard cases.
Configurations differ only in how much additional ground they fail to cover: $10$ residuals for GPT-5.5~(mid) in \modecodeproof, $11$ for GPT-5.5~(xhigh), and $28$ for Claude Sonnet~5, each a superset of the same seven.

\texttt{piggybank} shows the same structure with no exceptions at all: all eight configurations pass $17$ of $23$ specifications and all eight fail the same six.
Here the boundary is easy to characterise.
The $17$ that pass state properties of a single contract call, such as that inserting a negative amount fails or that smashing by a non-owner fails, and they reduce to case analysis on one invocation.
The six that fail quantify over reachable chain states or over an execution trace:

\begin{figure*}
\begin{lstlisting}[language=lean]
def spec_balance_on_chain (impl : RepoImpl) : Prop :=
  forall (bstate : ChainState) (caddr : Address),
    reachable bstate ->
      env_contracts bstate caddr = some (toWeakContract (contract impl)) ->
        outgoing_acts bstate caddr = [] ->
          exists cstate : State,
            contract_state bstate caddr = some cstate /\
              env_account_balances bstate caddr = cstate.balance
\end{lstlisting}
\end{figure*}

Discharging this requires induction over the reachability relation with a contract invariant that no single call establishes, which is the inductive-generalization gap of~\Cref{app:case:deposit} appearing as a hard boundary rather than a budget limit.
The division is exact: all six failing specifications quantify over a \texttt{ChainState} or a \texttt{ChainTrace}, and not one of the $17$ passing specifications does.
A single syntactic feature of the statement therefore predicts the outcome for every specification in the instance and for every configuration.

\texttt{unicode} localises the same phenomenon onto a semantic type.
Claude Opus~4.8 in \modeproof passes $40$ of $46$, and five of its six failures are encode and decode round-trips (\texttt{spec\_utf16\_roundtrip}, \texttt{spec\_utf16\_decode\_encode}, \texttt{spec\_uniChar\_utf16\_roundtrip}, \texttt{spec\_noChar\_utf8\_roundtrip}, and \texttt{spec\_noChar\_utf8\_decode\_encode}) while six other round-trip specifications in the same instance pass, among them \texttt{spec\_utf8\_roundtrip} and \texttt{spec\_uniChar\_utf8\_roundtrip}.
The split follows the level of abstraction rather than the round-trip property itself.
All three round-trips stated at the encoding-form level pass, including \texttt{spec\_utf16\_encode\_decode\_roundtrip}, so UTF-16 is not the obstacle.
Every failure is at the string level, where all three UTF-16 round-trips fail while the UTF-8 ones pass except in the variant restricted to strings without unicode characters.
The statements are near-identical in shape, each asserting that a checked encode followed by a checked decode recovers the input, so the difficulty sits in the supplied string-level conversion functions the proofs must unfold rather than in what the specifications say.

\subsection{Implementation Freedom Helps Only the Strongest Agent}
\label{app:mode-by-model}

The main text reports that \modecodeproof and \modeproof each win instances the other loses.
Disaggregating by model shows the effect is not symmetric across agents.
Counting instances where exactly one mode is a full solve, GPT-5.5~(xhigh) is \modecodeproof-only on $8$ and \modeproof-only on $6$. Every weaker configuration leans the other way, with GPT-5.5~(mid) at $2$ against $6$, Claude Opus~4.8 at $2$ against $4$, and Claude Sonnet~5 at $1$ against $1$.
At specification granularity, \modecodeproof gains GPT-5.5~(xhigh) $42$ specifications and GPT-5.5~(mid) $130$, but costs Claude Opus~4.8 $360$.

The cost is not proof difficulty.
Of Opus's $1{,}095$ residual specifications in \modecodeproof, $1{,}074$ are unattempted and only $21$ are rejected for axiom use, and none fails to build. In \modeproof its residual falls to $735$ and is entirely unattempted.
The trajectory data explain why.
Claude Opus~4.8 authors a median of $6$ implementation and $177$ proof lines by minute~$2$ of a \modecodeproof run and is still at $195$ proof lines by minute~$30$, reaching $811$ only in the final half hour.
GPT-5.5~(xhigh) already has $45$ implementation and $833$ proof lines at minute~$2$ and adds only $474$ more over the remaining $88$ minutes.
Opus therefore spends most of its wall clock before its proof text exists, so an added implementation obligation delays the part of the run that scores.
Six instances where its \modecodeproof score collapses to zero (\texttt{Huffman}, \texttt{ntheory}, \texttt{unicode}, \texttt{base58}, \texttt{dijkstar}, and \texttt{intervaltree}) score $8$ to $116$ in \modeproof, and in every one of its eight zero cells the grader finds no filled route at all rather than a failed one.

In short, \modecodeproof measures two capabilities at once, and for agents that are slow to produce a first artifact the implementation obligation dominates.
This is worth separating from the substitution effect of~\Cref{app:case:implementation-substitution}, which requires enough budget to first write an implementation and then exploit it.

\subsection{Weaker Agents Add Nothing to the Frontier}
\label{app:union}

A benchmark can be nearly saturated by an ensemble even when no single agent saturates it, so we ask how much the four agents cover together.

At repository granularity the answer is that they cover no more than the strongest one.
GPT-5.5~(xhigh) fully solves $33$ distinct instances across its two modes.
The union over all eight configurations is also $33$: not one instance is completed by a weaker configuration that GPT-5.5~(xhigh) misses in both modes.
The three weaker agents together solve $15$ instances, every one of which is a subset of what GPT-5.5~(xhigh) already solves.
Ensembling therefore buys nothing at the repository level, and the ten instances that no configuration completes resist the entire evaluation rather than resisting individual agents.

At specification granularity there is modest headroom.
Taking the union of every specification passed by any configuration gives $2{,}486$ of $2{,}705$ ($91.9\%$), against $2{,}362$ ($87.3\%$) for the best single cell.
The extra $124$ specifications are spread across instances where different agents happen to close different obligations, and they leave $219$ specifications that no configuration ever passes.
Those $219$ concentrate sharply, and every one of them falls in the ten instances that no configuration completes: $82$ in \texttt{dedekind\_reals}, where nothing is ever proved, then $34$ in \texttt{flocq}, $21$ each in \texttt{greenery} and \texttt{reedsolo}, $18$ in \texttt{galoistools}, $13$ in \texttt{ntheory}, $11$ in \texttt{base58}, $7$ in \texttt{verdict}, and $6$ each in \texttt{unicode} and \texttt{piggybank}.
This set is the part of \benchmark{} that is currently out of reach for every agent we tested, and it is small enough to inspect by hand, which is how we identified the reachability and abstraction-level patterns of~\Cref{app:walls}.
We further note that a repository where its implementation only satisfies a partial set of specifications is not considered verified.
Therefore, the union coverage at the specification level does not mean these specifications are saturated.

\subsection{Cost of Repository-Level Verification}
\label{app:cost}

\Cref{tab:cost} reports what each configuration spent in total.
Normalizing by what that spending bought gives a different ranking, shown in~\Cref{tab:cost-efficiency}.
These figures need a provenance caveat.
Costs are measured from the run's own telemetry for $121$ of the $344$ cells.
The other $223$ are runs that hit the wall-clock cap and were killed mid-turn before emitting a terminal usage record, so they are estimated as the model's median dollars per minute over its measured runs times that cell's elapsed time.
The estimated share is similar for every configuration, between $53\%$ and $66\%$ of total spend, so the comparison across configurations is at least uniform in method, but the absolute totals are soft and we read only the large gaps below.

\begin{table}[ht]
\centering
\caption{Cost per unit of progress. Total spend is summed over the $43$ instances, and the last two columns divide it by full solves and by passed specifications respectively.}
\label{tab:cost-efficiency}
\small
\setlength{\tabcolsep}{5pt}
\begin{tabular}{l l r r r r r}
\toprule
Agent & Mode & Total & Full & \$ / full solve & Specs & \$ / spec \\
\midrule
GPT-5.5 (xhigh)  & code + proof & \$2{,}865 & $27$ & \$106 & $2{,}362$ & \$1.21 \\
GPT-5.5 (xhigh)  & proof only   & \$2{,}964 & $25$ & \$119 & $2{,}320$ & \$1.28 \\
GPT-5.5 (medium) & code + proof & \$928     & $2$  & \$464 & $1{,}759$ & \$0.53 \\
GPT-5.5 (medium) & proof only   & \$1{,}094 & $6$  & \$182 & $1{,}629$ & \$0.67 \\
Claude Opus 4.8  & code + proof & \$1{,}983 & $8$  & \$248 & $1{,}610$ & \$1.23 \\
Claude Opus 4.8  & proof only   & \$2{,}279 & $10$ & \$228 & $1{,}970$ & \$1.16 \\
Claude Sonnet 5  & code + proof & \$633     & $2$  & \$317 & $1{,}270$ & \$0.50 \\
Claude Sonnet 5  & proof only   & \$791     & $2$  & \$396 & $1{,}237$ & \$0.64 \\
\bottomrule
\end{tabular}
\end{table}

\paragraph{The most capable configuration is also the cheapest per completed repository.}
GPT-5.5~(xhigh) spends the most in absolute terms and the least per full solve, at \$106 in \modecodeproof against \$182 to \$464 for every other configuration.
The ordering reverses per specification, where the two weakest configurations are the cheapest at \$0.50 and \$0.53, because partial coverage accumulates on easy obligations at low cost.
Which metric to prefer depends on what a practitioner wants.
Paying for specifications passed favors cheap agents, and paying for repositories actually completed favors the expensive one by a factor of two to four.

\paragraph{Raising reasoning effort is superlinear in yield.}
Moving GPT-5.5 from default to \texttt{xhigh} effort multiplies spend by $3.1$ in \modecodeproof and $2.7$ in \modeproof, and multiplies full solves by $13.5$ and $4.2$.
The gain is therefore larger than the price for this task, which follows from the all-or-nothing metric rather than from any superlinear gain in proof ability.
Per specification the same effort increase buys far less, moving coverage from $65\%$ to $87\%$ in \modecodeproof for triple the spend.
A configuration that closes most of a repository and stops earns nothing on the full-solve metric, so effort spent near the completion boundary converts into repository counts at a rate the specification counts do not show.

\paragraph{Unfinished runs cost more than finished ones.}
For three of the four agents the median unfinished cell is more expensive than the median full solve, at \$53 against \$43 for GPT-5.5~(xhigh), \$53 against \$42 for Claude Opus~4.8, and \$15 against \$9 for Claude Sonnet~5.
GPT-5.5~(mid) is the exception at \$16 against \$25, and the Sonnet and GPT-5.5~(mid) comparisons rest on four and eight full solves respectively, so we read the direction rather than the magnitudes.
Agents that finish tend to finish early, and agents that do not finish spend the remaining budget on obligations they never close.
Aggregated over the grid, $23\%$ of all spending went to the ten instances that no configuration solves.
This is the cost signature of the full-solve metric, and it is worth stating for anyone budgeting a run: much of the money is spent on the repositories that return nothing.

\clearpage

\section{Audit-Mechanism Case Studies}
\label{app:audit}
 
Each clean closure of an audit theorem constitutes formal evidence that a specification, as introduced in~\Cref{sec:benchmark:audit}, does not hold on the canonical implementation (\texttt{disprove}), is universally unsatisfiable (\texttt{unsat}), or jointly contradicts a sibling specification (\texttt{joint\_unsat}).
We ran the mechanism throughout curation, adjudicating every closure by hand against the pinned upstream source.
Restricted to the $43$ released instances, it produced $38$ adjudicated specification defects across $9$ instances, plus $6$ joint-unsatisfiability groups.
All of them are repaired in the released benchmark, and the evaluation reported in~\Cref{sec:eval} runs against the repaired version.
This appendix reports the aggregate findings first and then three case studies, one for each closure route: a \texttt{joint\_unsat} pair on \texttt{verdict}, an \texttt{unsat} family on \texttt{verified\_ironkv}, and the largest \texttt{disprove} cluster, on \texttt{verified\_bitmasks}.

A closure is only evidence if it is machine-checked, so we record the basis for each of the $38$ cases separately: $34$ rest on a clean certificate the agent submitted, $3$ on a certificate we reconstructed after the agent's proof failed the axiom check, and $1$ on an independently reproduced native evaluation.
Cases that share a root cause are not statistically independent, and we note the largest such cluster below rather than treating all $38$ as separate discoveries.
We also hold a further set of translated instances that we audited but did not bring to release quality, and their findings are excluded from every count here.
Those instances are the main reason the released corpus is $43$ rather than larger.

\begin{table}[h]
\centering
\caption{Audit-mechanism findings during benchmark development. ``Specs'' is the number of distinct specifications flagged across all configurations.}
\label{tab:audit-summary}
\small
\setlength{\tabcolsep}{4pt}
\begin{tabular}{l r r r l}
\toprule
Defect category & Cases & Share & Instances & Largest contributor \\
\midrule
Mistranslated definition or semantics   & $18$ & $47.4\%$ & $3$ & \texttt{huffman} ($16$) \\
Missing domain or operation condition   & $17$ & $44.7\%$ & $6$ & \texttt{verified\_bitmasks} ($7$) \\
Reference implementation mismatch       & $2$  & $5.3\%$  & $1$ & \texttt{flocq} ($2$) \\
Direct conflict between specifications  & $1$  & $2.6\%$  & $1$ & \texttt{verdict} ($1$) \\
\midrule
\textbf{Total}                          & \textbf{$38$} & & \textbf{$9$} & \\
\bottomrule
\end{tabular}
\end{table}

\paragraph{A second auditor adds little.}
We ran the audit with two agents, and their findings overlap heavily.
Claude Opus~4.8 found $33$ of the $38$ cases and Codex found $30$, with $25$ found by both.
Adding the weaker finder to the stronger one therefore contributes $5$ additional cases, a $15\%$ increment over the stronger single auditor.
This is useful for calibration: the marginal value of another auditor is real but modest, and the cases both agents find independently are the ones we adjudicated fastest, since two unrelated counterexamples for one specification leave little doubt about the verdict.

\paragraph{Audit~1: two specifications in direct conflict on \texttt{verdict} (\texttt{joint\_unsat}, fixed).}
Two specifications govern the Base64 character classifier \texttt{charToBits}:
 
\begin{lstlisting}[language=lean, basicstyle=\ttfamily\footnotesize]
def spec_char_to_bits_padding (impl : RepoImpl) : Prop :=
  impl.verdict.charToBits 0x3D = some none
 
def spec_char_to_bits_reject (impl : RepoImpl) : Prop :=
  ∀ (c : UInt8),
    (c < 0x2B ∨ (c > 0x2B ∧ c < 0x2F) ∨
     (c > 0x39 ∧ c < 0x41) ∨ (c > 0x5A ∧ c < 0x61) ∨ c > 0x7A) →
    impl.verdict.charToBits c = none
\end{lstlisting}
 
The padding character \texttt{`='} (\texttt{0x3D}) lies in the gap $(0\text{x}39, 0\text{x}41)$ between \texttt{`9'} and \texttt{`A'}, so \texttt{spec\_char\_to\_bits\_reject} requires \texttt{charToBits 0x3D = none} while \texttt{spec\_char\_to\_bits\_padding} requires it to return \texttt{some none}.
No implementation can satisfy both.
GPT-5.5~(xhigh) flagged this pair through a joint-unsatisfiability proof that collapses to a one-line $0\text{x}3\text{D} < 0\text{x}41 \land 0\text{x}3\text{D} > 0\text{x}39$ contradiction after unfolding both specifications.
The fix excludes \texttt{0x3D} from the reject range by adding a $\texttt{c} \ne \texttt{0x3D}$ conjunct:
 
\begin{lstlisting}[language=lean, basicstyle=\ttfamily\footnotesize]
     (c > 0x39 ∧ c < 0x41 ∧ c ≠ 0x3D) ∨    -- excluded `='
\end{lstlisting}
 
This error survived manual review and type-checking but was caught automatically by the agent.
 
\paragraph{Audit~2: comparator-lawfulness gap on \texttt{verified\_ironkv} (\texttt{unsat}, fixed).}
Three specifications on \texttt{verified\_ironkv} quantify over an arbitrary \texttt{KeyTrait~K} without constraining comparator lawfulness.
GPT-5.5~(xhigh) closed all three via \texttt{unsat} proofs by instantiating an adversarial comparator.
For example, the witness for \texttt{spec\_strictly\_ordered\_vec\_no\_duplicates} uses a constant comparator:
 
\begin{lstlisting}[language=lean, basicstyle=\ttfamily\footnotesize]
theorem unsat_strictly_ordered_vec_no_duplicates :
    ¬ (∃ impl : RepoImpl,
         spec_strictly_ordered_vec_no_duplicates impl) := by
  intro ⟨impl, h⟩
  -- Adversarial KeyTrait: cmpSpec returns Less for every pair.
  let bad : KeyTrait Unit :=
    { zeroSpec := (), cmpSpec := fun _ _ => Ordering.Less }
  -- A two-element vector of () is valid under bad but has duplicates.
  let s : StrictlyOrderedVec Unit := ⟨[(), ()]⟩
  exact (h _ bad s (by intro i j _; rfl)) ⟨0, 1⟩ rfl rfl
\end{lstlisting}
 
We resolved this by introducing a \texttt{LawfulKeyTrait} class capturing the minimal comparator laws (equality iff \texttt{.Equal}) and threading it through the three affected specifications.
 
\paragraph{Audit~3: missing domain conditions on \texttt{verified\_bitmasks} (\texttt{disprove}, fixed).}
The largest cluster of \texttt{disprove} closures in a released instance is $7$ specifications on \texttt{verified\_bitmasks}, and it separates into two root causes that both amount to a missing domain condition.

Three specifications assert that a bitwise operation preserves the bit count of its first argument, but they quantify over bitmasks of unequal length.
Because the implementation combines the arguments with \texttt{zipWith}, the result has length \texttt{min |A| |B|}, so any pair with \texttt{|B| < |A|} refutes the claim.
The repair adds the length precondition that the sibling \texttt{bSeq} specifications already carried:

\begin{lstlisting}[language=lean, basicstyle=\ttfamily\footnotesize]
def spec_bIIF_and_nbits_preserved (impl : RepoImpl) : Prop :=
  forall (A B : T), A.length = B.length ->        -- added
    impl.verifiedBitmasks.bIIF_nbits (impl.verifiedBitmasks.bIIF_and A B) =
      impl.verifiedBitmasks.bIIF_nbits A
\end{lstlisting}

The other four specifications compare a population count against a machine-word bound, and they fail once a bitmask is long enough for the count to exceed what a $64$-bit word represents.
This case is also the clearest example of why every closure passes through human adjudication.
Our first repair draft compared the counts in \texttt{Nat} instead of in the machine type, which is itself unsound because the residues still invert at lengths beyond $2^{64}$.
The second auditor rejected that draft and we adopted a uniform representability precondition for all four specifications instead.

After repair we did not merely re-run the audit to confirm the specifications are no longer refutable.
We wrote and compiled a positive certificate \texttt{theorem \_chk\_X : spec\_X canonical} for each of the $7$, so the record shows they are genuinely true of the reference implementation rather than only beyond the reach of the previous counterexample.

\clearpage

\section{Implementation Freedom in \modecodeproof Mode}
\label{app:case:implementation-substitution}

\modecodeproof mode lets an agent choose the implementation it will prove against, and the strongest agents exploit this by replacing a reference algorithm that is hard to reason about with one that is easy to reason about.
We audited the source of every \modecodeproof full solve whose \modeproof counterpart is not a full solve and identified at least five instance--agent pairs where the submitted implementation is algorithmically different from the reference rather than a stylistic rewrite.
Across those five, the agent closes all $250$ specifications in \modecodeproof, while proving the identical specifications against the fixed reference in \modeproof closes only $201$.

\paragraph{Replacing an algorithm with a simpler one.}
\texttt{munkres} is the clearest case, and all three of GPT-5.5~(xhigh), GPT-5.5~(mid), and Claude Opus~4.8 make the same move.
The reference implementation is a faithful port of the Hungarian algorithm: a $409$-line six-state machine driven by a fuel-bounded \texttt{runSteps} loop with $38$ auxiliary definitions covering starring, priming, and covering-line bookkeeping.
Reasoning about its output requires an invariant relating the internal state to assignment optimality at every step.
Instead, each agent submits a $164$-line implementation that enumerates permutations and selects a minimum-cost one:

\begin{lstlisting}[language=lean, basicstyle=\ttfamily\footnotesize]
def permsAuxImpl : List Nat -> List (List Nat)
  | [] => [[]]
  | x :: xs => (permsAuxImpl xs).flatMap (fun p => insertEverywhereImpl x p)

def permCostImpl (M : List (List Nat)) (p : List Nat) : Nat :=
  sumN ((List.range p.length).map (fun i => matGetN M i (natGet p i)))

def bestPermFrom (M : List (List Nat)) (best : List Nat) : List (List Nat) -> List Nat
  | [] => best
  | p :: ps =>
      if permCostImpl M p < permCostImpl M best then bestPermFrom M p ps
      else bestPermFrom M best ps
\end{lstlisting}

This implementation is exponential and would be unusable in production, but it is close to a transcription of what the specifications assert, so optimality follows by induction over the candidate list rather than from a loop invariant.
The mode gap is entirely in those obligations: in \modeproof every configuration closes exactly $12$ of $19$ specifications, and the seven failures are precisely the optimality and global-bound claims (\texttt{spec\_compute\_optimal\_value}, \texttt{spec\_compute\_achieves\_min}, \texttt{spec\_compute\_feasible}, \texttt{spec\_compute\_profit\_max}, \texttt{spec\_compute\_cost\_colVec}, \texttt{spec\_compute\_colVec\_mem}, and \texttt{spec\_compute\_cost\_global\_bounds}).
The structural specifications, which do not depend on the search strategy, pass in both modes.

\paragraph{Delegating to the standard library.}
The other two audited pairs substitute a library routine for a hand-written one.
On \texttt{sequences}, GPT-5.5~(xhigh) implements the merge-sort API as \texttt{fun a lessThanOrEq \_h => a.mergeSort lessThanOrEq}, and on \texttt{linked\_list} Claude Sonnet~5 implements list merging as \texttt{fun l => l.mergeSort (· $\leq$ ·)}.
Both replace a recursive definition the curator wrote with Lean's \texttt{List.mergeSort}, whose sortedness and permutation lemmas already exist in the standard library.
The effect is visible in the \texttt{linked\_list} pair, where Sonnet~5 closes $109$ of $109$ specifications in \modecodeproof but $81$ in \modeproof: the same proofs are available in both modes, but only in \modecodeproof can the agent choose the definition those proofs are stated about.

\paragraph{When implementation freedom hurts.}
The converse effect is larger across the corpus.
Seventeen instance--agent pairs are full solves in \modeproof but not in \modecodeproof, and the residuals there are dominated by unattempted obligations rather than rejected proofs.
\texttt{Huffman} with Claude Opus~4.8 is one of the example: $116$ of $127$ specifications in \modeproof versus $0$ in \modecodeproof.
The cause is not proof difficulty but the re-render boundary of~\Cref{app:anticheat}.
The agent left three implementation slots in \texttt{Impl/Build.lean} empty, so the graded artifact instantiates them with \texttt{sorry} where a function is required, and the build fails:

\begin{lstlisting}[basicstyle=\ttfamily\footnotesize]
error: Huffman/Impl/Build.lean:69:32: Function expected at
  ObuildfSig
but this term has type
  Type 1
\end{lstlisting}

Because \benchmark{} grades the repository rather than the specification, a single unbuildable implementation module voids every specification in the instance: the grader marks the submission \texttt{impl\_broken} and awards no partial credit.
Seven of the $344$ cells end this way, and all seven score zero.
This is the mechanism behind the mode asymmetry in~\Cref{fig:eval-modes}(a): \modeproof hands the agent a repository that already compiles, whereas \modecodeproof requires it to keep the repository compiling while it works.

\clearpage
\section{Hard-Case Proof Studies}
\label{app:case-studies}
\label{app:case:deposit}

The figures in~\Cref{sec:eval} report what agents achieve.
This appendix reports how they work, by reading the artifacts and the files agents leave behind in the sandbox.
We cover four instances:
\texttt{deposit\_sc} (the most substantive proof-engineering effort we observed),
\texttt{vest} (where case analysis migrates out of the obligations),
\texttt{sortedcontainers} (where the two task modes diverge sharply), and
\texttt{galoistools} (where an agent's search strategy is visible in its scratch files).

\subsection{Agents Leave Their Search Visible in the Sandbox}
\label{app:case-scratch}

Agents work in a sandbox containing the whole repository, and only marker interiors are graded.
Some agents exploit this freedom to build a private workspace: they create Lean files outside the graded regions, use them to probe the standard library and draft proofs, and leave them behind at the end of the run.
These files are not scored, but they record the agent's search in a way no summary statistic does.

The habit divides the agents sharply.
Claude Opus~4.8 creates such files in $32$ of its $86$ cells and Claude Sonnet~5 in $31$, together writing $360$ of them.
GPT-5.5~(xhigh) does so in $2$ cells and GPT-5.5~(mid) in $10$.
This is not a proxy for skill, and the direction is the opposite of what one might expect: for Opus, cells with a scratch workspace pass $59.0\%$ of specifications against $69.8\%$ for cells without one, and produce $2$ full solves against $16$.
Scratch files appear where an agent is stuck, so they are best read as a record of what stuck looks like.

\paragraph{Probing before proving.}
The first thing an agent does with a private file is ask what already exists.
Claude Sonnet~5 on \texttt{arithmetic} opens with a file that contains no proofs at all, only queries against the standard library:

\begin{lstlisting}[language=lean, basicstyle=\ttfamily\footnotesize]
import ArithmeticV2.Harness
import ArithmeticV2.Spec.DivMod

-- Test what lemmas exist
#check @Int.ediv_self
#check @Int.emod_lt_of_pos
#check @Int.ediv_add_emod
#check @Int.emod_emod_of_dvd
...
\end{lstlisting}

Later files in the same cell shift from probing to drafting, stating a private helper and a candidate obligation proof together so that both can be checked in one compilation before either is committed to a graded slot.
The agent has effectively built itself a read-eval-print loop out of files, and it is doing what a human would: establishing the available vocabulary first, then composing within it.
This behavior is invisible in the graded artifact, because the scratch files are discarded at grading time and the committed proofs show only the end state.

\paragraph{Retrying one lemma across many files.}
The same mechanism exposes a failure mode.
On \texttt{galoistools}, Claude Opus~4.8 in \modecodeproof leaves $76$ scratch files containing attempts at $42$ distinct lemmas, and the distribution is heavily skewed: $17$ lemmas are attempted in more than three files, and one, \texttt{refGfStrip\_eq\_gfStripAux}, reappears in $16$ of the $76$.
Three more (\texttt{refPolyEvalRevAux\_append\_zero}, \texttt{eval\_mod\_lemma}, and \texttt{cancel\_mod}) appear in $12$ or $13$ files each.
The cell closes $17$ of $48$ specifications.

What the agent is repeatedly trying to prove is a bridging lemma between the curator's reference definition and its own implementation:

\begin{lstlisting}[language=lean, basicstyle=\ttfamily\footnotesize]
-- Are refGfStrip and gfStripAux the same?
theorem refGfStrip_eq_gfStripAux : forall (f : List Nat),
    refGfStrip f = gfStripAux f := by
  intro f
  induction f with
  | nil => rfl
  | cons a as ih => simp [refGfStrip, gfStripAux]; split; exact ih; rfl
\end{lstlisting}

The agent has diagnosed the problem correctly.
Its own \texttt{gfStripAux} and the reference \texttt{refGfStrip} are extensionally equal, and the obligations need that fact, so the lemma is the right thing to want.
Across the sixteen files the statement stays fixed while the tactic script varies, and it also keeps returning to the lemma after intervening work on other goals rather than pursuing it to a conclusion in one sitting.

The alternative the agent never takes is to remove the need for the lemma.
Because this is \modecodeproof mode, it controls \texttt{gfStripAux} and could define it to be syntactically the reference function, at which point the agreement lemma becomes \texttt{rfl}.
This is precisely the substitution move that succeeds on \texttt{munkres} and \texttt{sortedcontainers} (\Cref{app:case:implementation-substitution,app:case-sortedcontainers}), applied in reverse: instead of choosing an implementation that is easy to reason about, the agent commits to one that differs from the reference and then pays for the difference in every proof.

\paragraph{Implication for agent design.}
Two properties of this behavior suggest concrete changes.
First, the probing phase is cheap and effective, and the agents that skip it are the ones that already know the library. An agent with weaker priors benefits from being asked to enumerate the available lemmas before drafting.
Second, the retry loop has no step that revisits the implementation.
The agent treats its own definitions as fixed once written, so a mismatch it introduced in the first minutes becomes a proof obligation it pays for over the remaining ninety.
This is the same one-way ordering that~\Cref{fig:eval-dynamics} shows across all configurations, where implementation size plateaus early while proof text grows to the deadline.
A loop that, on repeated failure of a bridging lemma, reconsidered the definition rather than the tactic would turn sixteen attempts at one lemma into one edit that discharges it.
 
\subsection{\texttt{deposit\_sc}: A Layered Proof Library}
\label{app:case-deposit-sc}
 
\texttt{deposit\_sc} is a Lean port of the ConsenSys Dafny Merkle-tree incremental-commitment contract; the instance contains $79$ specifications across four modules (\texttt{Bits}, \texttt{Tree}, \texttt{Merkle}, \texttt{Contract}).
GPT-5.5~(xhigh) closes all $79$ in both task modes.
The \modecodeproof submission is the largest proof artifact in the corpus: $6{,}052$ nonblank lines across the four proof modules, containing $99$ agent-written helper theorems alongside the $79$ specification solutions.
The library is layered to match the module hierarchy, with bit-list arithmetic (\texttt{Bits}, $846$ lines), structural facts about complete decorated trees (\texttt{Tree}, $1{,}384$), \texttt{buildMerkle} and root-computation correctness (\texttt{Merkle}, $3{,}479$), and state-transformer frame conditions (\texttt{Contract}, $343$), so that each tier depends only on the ones below it.

\paragraph{The work happens in named helpers.}
$54.3\%$ of the proof text sits in helper theorems rather than in the obligations themselves, and the two populations have very different shapes: the median obligation proof is $9$ lines while the median helper is $19$, and the longest helper runs $262$ lines.
The effect is that most obligations reduce to a single application.
\texttt{prove\_same\_leaves\_same\_root} states that two complete decorated trees with the same height and leaf sequence have equal roots, and its proof is a two-line delegation to a $94$-line induction the agent proved separately:

\begin{lstlisting}[language=lean, basicstyle=\ttfamily\footnotesize]
theorem prove_same_leaves_same_root : spec_same_leaves_same_root canonical := by
  intro t t' f hc hc' hd hd' hh hleaves
  exact tree_same_leaves_same_root_aux t t' f hc hc' hd hd' hh hleaves
\end{lstlisting}

This is the division of labor the aggregate statistics in~\Cref{fig:eval-proof-architecture} describe, visible in one instance: the agent decides what the reusable fact is, proves it once under a name, and then discharges specifications by citing it.

\paragraph{Strengthening the induction hypothesis.}
The interesting proofs are the ones where the obligation as stated is not directly inductive.
\texttt{spec\_incremental\_equals\_batch} asserts that folding a batch of deposits agrees with depositing them one at a time.
Proving it by induction on the batch fails, because the inductive step needs to know that the intermediate state is still valid and that its height, merge function, and default are unchanged.
The agent's proof introduces a conjunctive auxiliary claim carrying all five facts at once and induct on that:

\begin{figure*}
\begin{lstlisting}[language=lean, basicstyle=\ttfamily\footnotesize]
have aux : forall (s : DepositState) (vs : List Int),
    Valid canonical s →
    s.values.length + vs.length < canonical.depositSc.power2 s.treeHeight →
    Valid canonical (vs.foldl DepositSc.deposit s) ∧
    (vs.foldl DepositSc.deposit s).values = s.values ++ vs ∧
    (vs.foldl DepositSc.deposit s).treeHeight = s.treeHeight ∧
    (vs.foldl DepositSc.deposit s).f = s.f ∧
    (vs.foldl DepositSc.deposit s).default = s.default := by
  intro s vs
  induction vs generalizing s with
  ...
\end{lstlisting}
\end{figure*}

Choosing that conjunction is the step our depth analysis identifies as the discriminator between configurations: the fact needed at depth four is not a harder version of the fact needed at depth zero, it is a differently stated one.

\paragraph{What changed relative to earlier evaluations.}
An earlier round of this evaluation left $7$ of the $79$ specifications unproven, all depending on one missing boundary-shift lemma about how Merkle-path siblings evolve when the leaf array grows by one element. Those seven included \texttt{spec\_deposit\_preserves\_valid}, the contract's principal inductive invariant.
In the current evaluation the agent closes that invariant, and its proof is where the remaining difficulty concentrated: a $36$-line argument that destructures the eight-way validity conjunction, rebuilds the sibling relation for the extended leaf array, and re-establishes each component.
The longest single proof in the submission is now \texttt{prove\_compute\_new\_left\_path\_correct} at $524$ lines.
The limitation we described previously was therefore a budget and organization boundary rather than a capability ceiling: given the same specifications, the same configuration eventually found the generalization it had missed.

\subsection{\texttt{vest}: Parser Branch Reasoning}
\label{app:case-vest}
 
\texttt{vest} is a Lean port of selected modules from the \texttt{secure-foundations/vest} verified-parser project, with $42$ specifications spread thinly across $13$ modules, none of which carries more than eight.
GPT-5.5~(xhigh) and Claude Opus~4.8 close all $42$ in both modes, GPT-5.5~(mid) closes $40$ and $42$, and Claude Sonnet~5 closes $33$ and $38$.
The instance is therefore useful for a different question than \texttt{deposit\_sc}: not whether an agent can sustain a long proof, but where the work goes when the specifications are shallow and numerous.

\paragraph{Case analysis migrates into helpers.}
The obligations themselves are short.
\texttt{spec\_btcvarint\_parse\_correct}, which relates the executable Bitcoin variable-length-integer parser to its specification, closes in six tactic lines:

\begin{lstlisting}[language=lean, basicstyle=\ttfamily\footnotesize]
theorem prove_btcvarint_parse_correct : spec_btcvarint_parse_correct canonical := by
  intro s
  constructor
  . intro n v h
    simp [canonical, VestV2.btcVarintParse, h]
  . intro h
    exact <ParseError.UnexpectedEndOfInput,
           by simp [canonical, VestV2.btcVarintParse, h]>
\end{lstlisting}

The four-way case analysis the format actually requires has not disappeared. It has moved into a named helper.
The submission's longest proof is an $87$-line bound on the parser's consumed length, which splits on the leading discriminant byte to separate the single-byte, \texttt{0xFD}, \texttt{0xFE}, and \texttt{0xFF} encodings and then destructures enough of the payload in each branch to establish the bound:

\begin{figure*}
\begin{lstlisting}[language=lean, basicstyle=\ttfamily\footnotesize]
theorem btc_parse_length_bounds_aux :
    forall (s : List UInt8) (n : Int) (v : VarInt),
      VarInt.spec_parse s = some (n, v) ->
      1 <= n /\ n <= 9 /\ n.toNat <= s.length := by
  intro s n v h
  unfold VarInt.spec_parse at h
  cases s with
  | nil => simp at h
  | cons b rest =>
      by_cases hb : b.toNat <= 0xFC
      . simp [hb] at h; rcases h with <rfl, rfl>; simp
      . simp [hb] at h
        by_cases hfd : b.toNat = 0xFD
        ...
\end{lstlisting}
\end{figure*}

Across the instance the agent writes $27$ helper theorems for $1{,}893$ lines of proof, roughly one helper per one-and-a-half specifications.
The contrast with \texttt{deposit\_sc}, where $99$ helpers support $79$ specifications over $6{,}052$ lines, is the difference between a repository whose difficulty is depth and one whose difficulty is breadth: \texttt{vest} needs many small facts about independent formats, and the agent's library is correspondingly flat.

\subsection{\texttt{sortedcontainers}: Choosing an Algorithm You Can Induct On}
\label{app:case-sortedcontainers}

For \texttt{sortedcontainers}, GPT-5.5~(xhigh) closes all $49$ specifications in \modecodeproof and $24$ in \modeproof, with the $25$ residuals distributed across all three data structures (\texttt{SortedDict} $11$, \texttt{SortedList} $7$, \texttt{SortedSet} $7$) and every one of them unattempted rather than rejected.
The instance is a Lean port of the \texttt{python-sortedcontainers} library, and its reference implementation follows the Python original in using binary search over an index range:

\begin{lstlisting}[language=lean, basicstyle=\ttfamily\footnotesize]
def bisectLeftGo (lst : List a) (val : a) (lo hi : Nat) : Nat :=
  ...
  if ... then bisectLeftGo lst val (mid + 1) hi
         else bisectLeftGo lst val lo mid
termination_by hi - lo
\end{lstlisting}

Proving anything about \texttt{bisectLeftGo} requires an invariant tying the \texttt{lo}/\texttt{hi} window to the sortedness of the enclosing list, maintained across a recursion whose measure is \texttt{hi - lo} rather than the structure of the list.
That invariant is what the $25$ unattempted \modeproof specifications need, and no configuration supplies it.

In \modecodeproof the agent avoids the invariant entirely by choosing a different algorithm.
It adds $110$ lines to \texttt{Impl/SortedList.lean} implementing insertion-based sorting and a linear scan in place of bisection:

\begin{lstlisting}[language=lean, basicstyle=\ttfamily\footnotesize]
def linearBisectLeft {a : Type} [Ord a] (value : a) : List a -> Nat
  | [] => 0
  | x :: xs => if ordLt x value then linearBisectLeft value xs + 1 else 0

def insertOrd {a : Type} [Ord a] (value : a) : List a -> List a
  ...
def sortOrd {a : Type} [Ord a] : List a -> List a
  | x :: xs => insertOrd x (sortOrd xs)
\end{lstlisting}

Both definitions recurse on the list constructor, so every specification about them is provable by \texttt{induction} on the list with no auxiliary window invariant.
The substitution costs asymptotic efficiency, giving $O(n)$ lookup instead of $O(\log n)$ and $O(n^2)$ construction instead of $O(n \log n)$, so the agent's implementation is not a drop-in replacement for the library it ports.
It nonetheless satisfies every scored specification, because the specifications constrain the input--output behavior of the container rather than its complexity.

This illustrates a limitation of our metric that~\Cref{app:anticheat} discusses in general terms.
A full solve in \modecodeproof certifies that the submitted implementation meets the submitted specifications, but it does not certify that the implementation preserves properties the specifications do not mention.
Complexity is the most consequential such property for a data-structure library, and it is exactly what an agent will trade away when the proof obligation is the binding constraint.
Specifying cost behavior formally is possible in principle but requires a cost model the source repositories do not provide, so we note the gap rather than close it.


%% file: figures/tab_per_instance.tex
\begin{table}[ht]
\centering
\caption{Per-instance pass counts (passed / total scored specifications) at each agent's final submitted state, across all evaluated cells for the 43 instances. ``cp'' = code-and-proof, ``po'' = proof-only.}
\label{tab:per-instance}
\small
\setlength{\tabcolsep}{2.5pt}
\resizebox{\textwidth}{!}{
\begin{tabular}{l rr rr rr rr}
\toprule
& \multicolumn{2}{c}{GPT-5.5 (xhigh)} & \multicolumn{2}{c}{GPT-5.5 (medium)} & \multicolumn{2}{c}{Claude Opus 4.8} & \multicolumn{2}{c}{Claude Sonnet 5} \\
\cmidrule(lr){2-3} \cmidrule(lr){4-5} \cmidrule(lr){6-7} \cmidrule(lr){8-9}
Instance & cp & po & cp & po & cp & po & cp & po \\
\midrule
\multicolumn{9}{l}{\emph{Track~1: Formal-source instances}} \\
\texttt{arithmetic} & 191/191 & 189/191 & 139/191 & 191/191 & 191/191 & 191/191 & 128/191 & 117/191 \\
\texttt{dedekind\_reals} & 0/82 & 0/82 & 0/82 & 0/82 & 0/82 & 0/82 & 0/82 & 0/82 \\
\texttt{deposit\_sc} & 79/79 & 79/79 & 63/79 & 65/79 & 71/79 & 74/79 & 21/79 & 48/79 \\
\texttt{flocq} & 155/203 & 154/203 & 138/203 & 150/203 & 153/203 & 166/203 & 119/203 & 148/203 \\
\texttt{huffman} & 127/127 & 92/127 & 14/127 & 19/127 & 0/127 & 116/127 & 42/127 & 72/127 \\
\texttt{json} & 64/64 & 61/64 & 61/64 & 64/64 & 42/64 & 64/64 & 50/64 & 35/64 \\
\texttt{piggybank} & 17/23 & 17/23 & 17/23 & 17/23 & 17/23 & 17/23 & 17/23 & 17/23 \\
\texttt{sequences} & 84/84 & 84/84 & 84/84 & 25/84 & 21/84 & 71/84 & 0/84 & 65/84 \\
\texttt{unicode} & 0/46 & 36/46 & 0/46 & 32/46 & 0/46 & 40/46 & 0/46 & 26/46 \\
\texttt{verdict} & 108/119 & 97/119 & 109/119 & 0/119 & 112/119 & 52/119 & 91/119 & 0/119 \\
\texttt{verified\_bitmasks} & 124/124 & 124/124 & 105/124 & 105/124 & 103/124 & 124/124 & 71/124 & 69/124 \\
\texttt{verified\_ironkv} & 19/22 & 22/22 & 17/22 & 14/22 & 19/22 & 19/22 & 11/22 & 14/22 \\
\texttt{vest} & 42/42 & 42/42 & 40/42 & 42/42 & 42/42 & 42/42 & 33/42 & 38/42 \\
\midrule
\multicolumn{9}{l}{\emph{Track~2: Python-source instances}} \\
\texttt{base58} & 25/53 & 20/53 & 19/53 & 19/53 & 0/53 & 35/53 & 6/53 & 11/53 \\
\texttt{cachetools} & 73/74 & 74/74 & 54/74 & 45/74 & 35/74 & 68/74 & 35/74 & 0/74 \\
\texttt{croniter} & 55/55 & 55/55 & 49/55 & 52/55 & 54/55 & 53/55 & 45/55 & 46/55 \\
\texttt{difflib} & 49/49 & 49/49 & 29/49 & 42/49 & 33/49 & 26/49 & 24/49 & 0/49 \\
\texttt{dijkstar} & 43/43 & 35/43 & 32/43 & 11/43 & 0/43 & 29/43 & 11/43 & 11/43 \\
\texttt{ecdsa} & 48/48 & 48/48 & 43/48 & 37/48 & 46/48 & 48/48 & 39/48 & 31/48 \\
\texttt{galoistools} & 30/48 & 14/48 & 16/48 & 14/48 & 17/48 & 20/48 & 15/48 & 20/48 \\
\texttt{greenery} & 5/26 & 2/26 & 5/26 & 2/26 & 5/26 & 2/26 & 5/26 & 2/26 \\
\texttt{intervaltree} & 56/56 & 55/56 & 34/56 & 16/56 & 0/56 & 8/56 & 8/56 & 13/56 \\
\texttt{ipaddress} & 68/68 & 68/68 & 49/68 & 68/68 & 44/68 & 48/68 & 21/68 & 42/68 \\
\texttt{jsonpatch} & 27/27 & 27/27 & 24/27 & 27/27 & 27/27 & 27/27 & 23/27 & 27/27 \\
\texttt{linked\_list} & 109/109 & 109/109 & 108/109 & 87/109 & 109/109 & 109/109 & 109/109 & 81/109 \\
\texttt{munkres} & 19/19 & 12/19 & 19/19 & 12/19 & 19/19 & 12/19 & 7/19 & 12/19 \\
\texttt{netaddr} & 40/40 & 40/40 & 23/40 & 23/40 & 24/40 & 24/40 & 14/40 & 18/40 \\
\texttt{networkx} & 40/40 & 40/40 & 24/40 & 24/40 & 39/40 & 22/40 & 20/40 & 20/40 \\
\texttt{ntheory} & 43/62 & 48/62 & 34/62 & 36/62 & 0/62 & 42/62 & 30/62 & 29/62 \\
\texttt{packaging\_version} & 70/74 & 74/74 & 69/74 & 31/74 & 22/74 & 51/74 & 26/74 & 0/74 \\
\texttt{portion} & 63/63 & 63/63 & 13/63 & 29/63 & 0/63 & 0/63 & 20/63 & 9/63 \\
\texttt{primefac} & 56/56 & 56/56 & 20/56 & 51/56 & 43/56 & 47/56 & 16/56 & 18/56 \\
\texttt{primepy} & 9/9 & 9/9 & 8/9 & 6/9 & 9/9 & 9/9 & 8/9 & 7/9 \\
\texttt{prolepticgregorian} & 61/61 & 52/61 & 58/61 & 0/61 & 32/61 & 38/61 & 15/61 & 18/61 \\
\texttt{pyradix} & 52/52 & 52/52 & 50/52 & 52/52 & 35/52 & 1/52 & 45/52 & 31/52 \\
\texttt{pythonconstraint} & 20/20 & 20/20 & 18/20 & 18/20 & 20/20 & 20/20 & 20/20 & 20/20 \\
\texttt{reedsolo} & 0/48 & 27/48 & 12/48 & 9/48 & 14/48 & 19/48 & 7/48 & 9/48 \\
\texttt{rsa} & 59/63 & 63/63 & 30/63 & 33/63 & 49/63 & 33/63 & 35/63 & 29/63 \\
\texttt{semver} & 52/53 & 53/53 & 52/53 & 52/53 & 48/53 & 53/53 & 35/53 & 22/53 \\
\texttt{sortedcontainers} & 49/49 & 24/49 & 15/49 & 18/49 & 33/49 & 40/49 & 14/49 & 11/49 \\
\texttt{textdistance} & 59/59 & 59/59 & 20/59 & 47/59 & 17/59 & 58/59 & 0/59 & 25/59 \\
\texttt{textwrap} & 60/60 & 60/60 & 39/60 & 39/60 & 50/60 & 42/60 & 29/60 & 20/60 \\
\texttt{toposort} & 12/15 & 15/15 & 6/15 & 5/15 & 15/15 & 10/15 & 5/15 & 6/15 \\
\bottomrule
\end{tabular}}
\end{table}